\documentclass[letterpaper, 10 pt, conference]{ieeeconf}

\usepackage{amsmath,amssymb,amsfonts,mathtools}
\usepackage{booktabs,multirow,tabularx,array}
\usepackage{tabularray}
\UseTblrLibrary{booktabs}
\usepackage{graphicx}
\usepackage{capt-of}
\usepackage{xcolor}
\usepackage{xspace}
\usepackage{placeins}
\usepackage{afterpage}
\makeatletter
\newcommand\PinFloat[3]{\@ifundefined{pin@#1@#2}{\global\@namedef{pin@#1@#2}{}}{}%
  \expandafter\g@addto@macro\csname pin@#1@#2\endcsname{#3}}
\newcommand\pin@col{\if@firstcolumn 1\else 2\fi}
\newcommand\pin@run{\@ifundefined{pin@\the\value{page}@\pin@col}{}{%
  \csname pin@\the\value{page}@\pin@col\endcsname\global\expandafter\let\csname pin@\the\value{page}@\pin@col\endcsname\relax}}
\newcommand\pin@check{\afterpage{\pin@run\pin@check}}
\AtBeginDocument{\@ifundefined{r@fig:pull}{\global\let\pin@run\relax}{}}
\makeatother
\usepackage{url}
\usepackage{cite}
\usepackage{hyperref}
\usepackage[figure,figure*,table,table*]{hypcap}
\makeatletter
\newcommand\floatcounter{\global\advance\csname c@\@captype\endcsname\@ne\global\@capstartfalse
  \global\let\@currentHref\hc@currentHref\protected@edef\@currentlabel{\csname the\@captype\endcsname}}
\makeatother

\IEEEoverridecommandlockouts

\newcommand{\system}{\textup{SimScale}\xspace}
\newcommand{\assetgen}{\textup{AssetEngine}\xspace}
\newcommand{\demogen}{\textup{DemoEngine}\xspace}

\title{\LARGE \bf
SimScale: Agentic Simulation Data Scaling \\ for Sim-to-Real Policy Learning\vspace{-3pt}}

\author{Tianyuan Dai$^{1,2*}$, Kevin Lin$^{1,2*}$, Yunfan Jiang$^{2}$, Minghuan Liu$^{1}$, Junhong Xu$^{1}$, Justin Tran$^{2}$,\\
Benjamin Nguyen$^{1}$, Yingrui Huang$^{1}$, Linxi ``Jim'' Fan$^{2\dagger}$, Ajay Mandlekar$^{2\dagger}$, Yuke Zhu$^{1,2\dagger}$\\
$^{1}$UT Austin\quad $^{2}$NVIDIA%
\thanks{$^{*}$Equal contribution. $^{\dagger}$Equal advising.}}

\begin{document}
\bstctlcite{IEEEexample:BSTcontrol}
\setlength{\skip\footins}{2pt plus 1pt}

\IEEEaftertitletext{\vspace{\dimexpr-7mm+9pt\relax}\noindent\begin{minipage}{\textwidth}
\centering
\providecommand{\teaserwidth}{\textwidth}
{\offinterlineskip
\includegraphics[width=\teaserwidth,trim=0 72.87 0 6,clip]{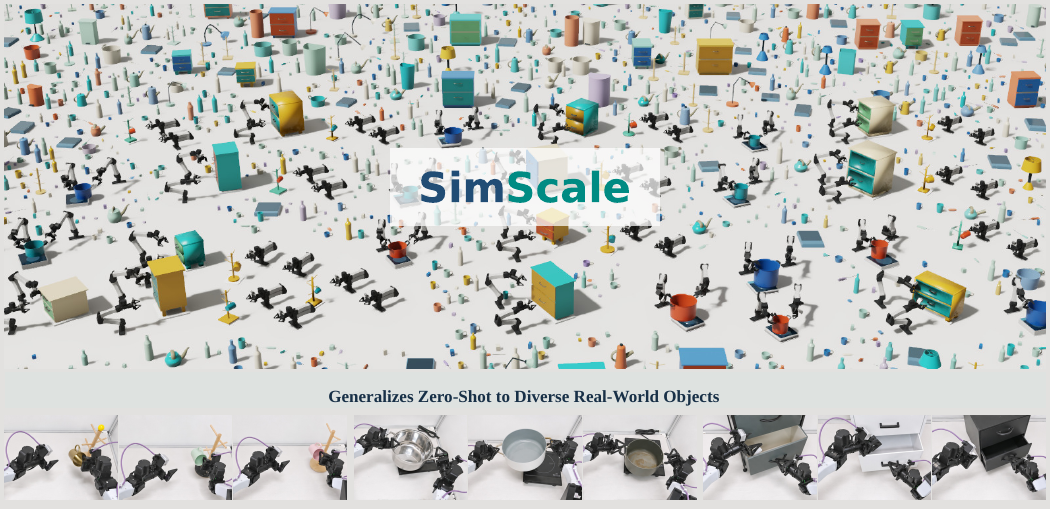}\par
\includegraphics[width=\teaserwidth,trim=0 4.37 0 183.5,clip]{figs/simscale_scale_teaser.pdf}\par}\vspace{-3pt}%
\captionof{figure}{\textbf{\system Overview.} \system scales up simulation data from one human demonstration in simulation by automatically defining diverse, physically plausible assets and generating demonstrations across their diverse geometries. RGB visuomotor policies generalize zero-shot to diverse real-world objects.}
\label{fig:pull}
\end{minipage}
\vspace{1.3em}}
\maketitle
\thispagestyle{empty}
\pagestyle{empty}

\PinFloat{5}{1}{\begin{figure}[t]
    \centering
    \includegraphics[width=\linewidth]{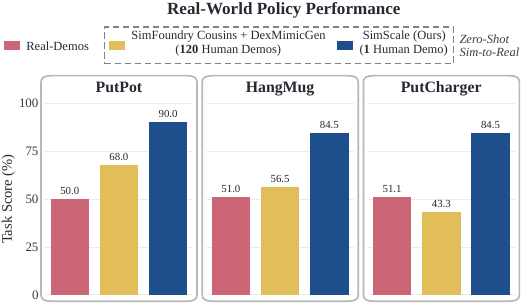}
    \vspace{-6mm}
    \caption{\textbf{\system policies generalize to diverse real-world objects with only one human demonstration and no real-world data.}
    Real-world policy performance is averaged over five object sets per task, with DINOv3 + FMT. GR00T N1.6 and other simulation baselines are in Figs.~\ref{fig:real-world-groot} and~\ref{fig:real-world-assets}.}
    \label{fig:real-world-compact}
    \vspace{-3mm}
\end{figure}
}
\PinFloat{5}{2}{\begin{figure}[t]
\centering
\includegraphics[width=\linewidth]{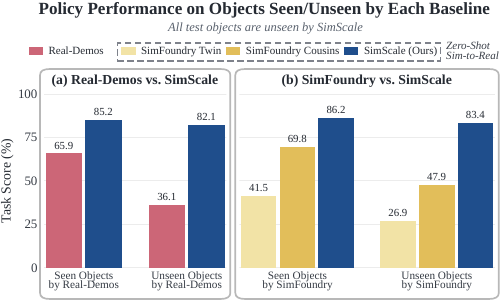}
\vspace{-6.5mm}
\caption{\textbf{\system generalizes to various unseen real-world objects, while baselines degrade sharply on unseen objects.}
\system never sees any test objects. Bars average three tasks and two policy backends, with per-task results in Figs.~\ref{fig:seen-unseen-real-demos} and~\ref{fig:seen-unseen-simfoundry}.}
\label{fig:real-generalization}
\vspace{-3mm}
\end{figure}
}
\PinFloat{6}{1}{\begin{figure}[t]
    \centering
    \includegraphics[width=\linewidth]{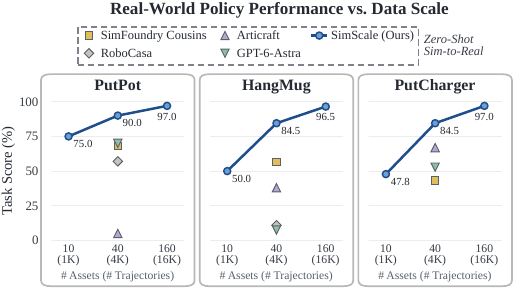}
    \vspace{-6mm}
    \caption{\textbf{\system policy performance scales with the number of synthetically generated assets and trajectories.}
    Policies are trained with DINOv3 + FMT on 100 trajectories per asset. With 10 assets and one human demonstration, \system outperforms on average all baselines with 40 assets and 120 human demonstrations, showing \textbf{much higher data quality}.}
    \label{fig:real-world-scaling}
    \vspace{-3mm}
\end{figure}
}
\PinFloat{6}{2}{\begin{figure}[t]
\centering
\includegraphics[width=\linewidth]{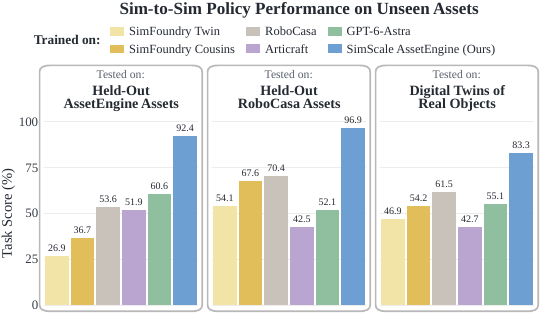}
\vspace{-6mm}
\caption{\textbf{Policies trained on assets generated by \system \assetgen exhibit stronger geometric generalization under sim-to-sim settings.}
Bars average three tasks with DINOv3 + FMT, with per-task results for both policy backends in Figs.~\ref{fig:sim-test-assetengine}--\ref{fig:sim-test-twins}.}
\label{fig:sim-assets-all}
\vspace{-3mm}
\end{figure}
}
\PinFloat{7}{1}{\begin{figure}[t]
\centering
\includegraphics[width=\linewidth]{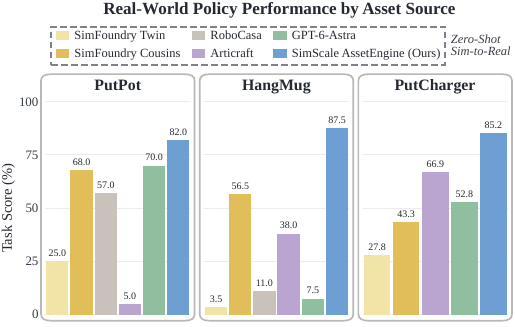}
\vspace{-6mm}
\caption{\textbf{Policies trained on assets generated by \system \assetgen show stronger zero-shot transfer to the real world.}
All policies are trained with DINOv3 + FMT. GR00T N1.6 results are in Fig.~\ref{fig:real-world-assets-groot}.}
\label{fig:real-world-assets}
\vspace{-3mm}
\end{figure}
}
\PinFloat{7}{2}{\begin{figure}[t]
    \centering
    \includegraphics[width=\linewidth]{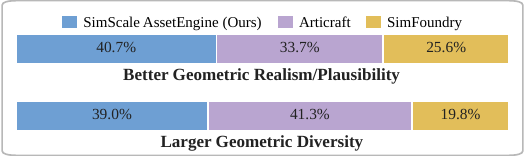}\\[-.5mm]
    {\footnotesize (a) User study on task-relevant objects}\\[1.5mm]
    \includegraphics[width=\linewidth]{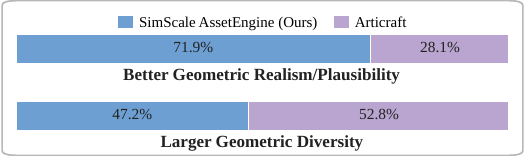}\\[-.5mm]
    {\footnotesize (b) User study on 20 categories including more complex objects}
    \vspace{-1mm}
    \caption{\textbf{Assets generated by \system \assetgen achieve the best balance between geometric diversity and plausibility based on human preference.}
    Participants rank assets from different generation methods in terms of geometric realism/plausibility and diversity, with results on task-relevant object categories in (a) and on 20 categories including more complex objects in (b).}
    \label{fig:asset-preference-all-categories}
    \vspace{-3mm}
\end{figure}
}
\PinFloat{8}{1}{\begin{figure}[t]
    \centering
    \includegraphics[width=\linewidth]{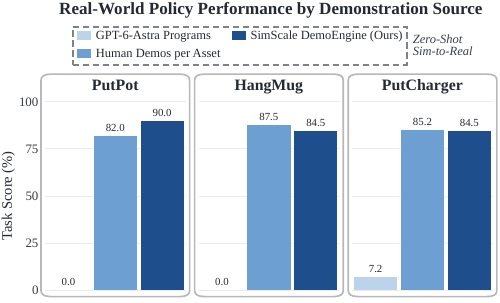}
    \vspace{-4.33mm}
    \caption{\textbf{Our \system \demogen generates data from one human demonstration on par with human demonstrations re-collected for every asset.}
    All methods are trained with DINOv3 + FMT on the same 40 \assetgen assets. GR00T N1.6 results are in Fig.~\ref{fig:real-world-demo-groot}.}
    \label{fig:real-world-demo}
    \vspace{-3mm}
\end{figure}
}
\makeatletter\pin@check\makeatother

\begin{abstract}
Simulation offers a scalable source of robot training data, but constructing diverse, physically plausible assets and generating demonstrations across their different geometries remain two major bottlenecks.
We introduce \system, an agentic synthetic data generation framework for policy generalization across diverse object geometries.
Given a short description of an object category, \system searches diverse internet images and distills their shared structures and geometric variations into a category-level procedural generator. The generator is iteratively revised until it passes independent reviews for category coverage and geometric validity, yielding diverse, physically plausible, and simulation-ready assets. Given one human demonstration in simulation, \system adapts it across the diverse geometric variations, scaling it into thousands of trajectories across geometries and poses.
RGB visuomotor policies trained on this data transfer zero-shot to diverse real-world objects.
Across more than 5,000 real-robot trials spanning three bimanual manipulation tasks and two policy architectures, \system outperforms policies trained on real-world demonstrations and on simulation data from existing asset libraries, real-to-sim methods, and agentic asset generation methods, by approximately 30\% on average over the strongest baseline.
Scaling \system data steadily improves real-world policy performance to more than 96\% in all three tasks without any real-world data, from 10 to 160 assets and from 1,000 to 16,000 trajectories per task, totaling over 300 hours of data automatically generated from one human demonstration per task.
Additional results are at \href{https://simscale-agentic.github.io/}{\textcolor{blue}{simscale-agentic.github.io}}.
\end{abstract}

\section{Introduction}
\label{sec:introduction}
\begin{figure*}[!t]
    \centering
    \includegraphics[width=\textwidth]{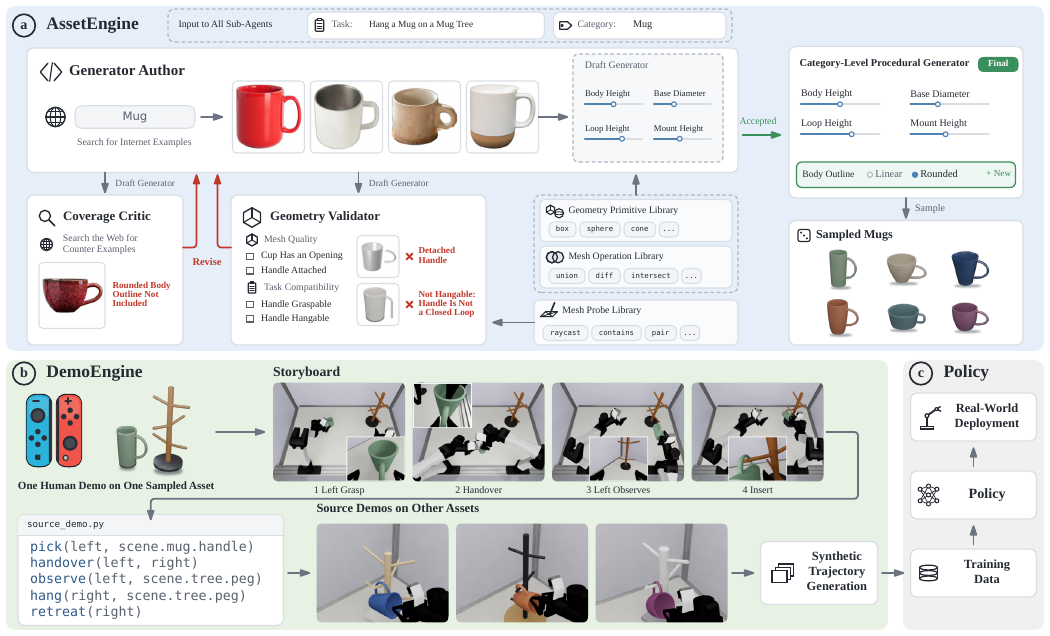}
    \vspace{-14.5pt}
    \caption{\textbf{\system Method.}
    \system scales one human demonstration into demonstrations on diverse and physically plausible
    assets with two agentic engines.
    (a) In \assetgen, the Generator Author defines a category-level procedural generator, grounded by internet
    images, to cover real-world variations. The Coverage Critic and Geometry Validator diagnose the geometry
    coverage and physical plausibility of sampled assets and send feedback to the Generator Author for
    revision. The procedural generator is finalized once the three sub-agents reach a consensus.
    (b) \demogen extracts a storyboard from the human demonstration and writes a source-demo script that
    adapts it to diverse object geometries. DexMimicGen further augments these source demonstrations across
    object poses.
    (c) RGB visuomotor policies are trained on the resulting data and deployed zero-shot on diverse unseen
    real objects.}
    \label{fig:agentic-method}
    \vspace{-4mm}
\end{figure*}

Generalization to geometrically diverse objects is a central challenge in robot learning.
Robot foundation models~\cite{brohan2023rt2,nvidia2025gr00t,nvidia2026gr00t2,black2024_0,black2025pi05,intelligence2026pi07,liu2024rdt,agibot2025colosseo,cheang2025gr3} show that training on large amounts of robot demonstrations benefits policy generalization across tasks and scenes.
Demonstrations that span a wide spectrum of geometric variations can help policies generalize to diverse unseen objects, but collecting them in the real world requires substantial human effort. Simulation offers a practical and cost-effective way to generate this diverse training data~\cite{mandlekar2023mimicgen,dai2024automated,deshpande2026molmobot,tian2025interndata,deng2025graspvla,yin2026omnireset}, but doing so requires constructing diverse and physically plausible assets, generating demonstrations across their geometries, and overcoming sim-to-real gaps.

Scaling simulation data across object geometries faces two coupled bottlenecks. First, assets must cover meaningful real-world variations while remaining physically plausible.
Curated asset libraries~\cite{li2022behaviork,nasiriany2024robocasa,iliash2026artiverse,jin2025artvip} collect handcrafted assets, which are physically plausible but cover few instances per category. Real-to-sim methods~\cite{torne2024rialto,dai2024automated,ranawaka2026simfoundry,han2025re3sim,li2026arstag,zhang2026autorsr} reconstruct realistic digital twins of observed objects and optionally augment their geometry by generating constrained variants, i.e., digital cousins, but these variants stay tied to a specific scene and lack coverage of the rest of the category. Agentic generators~\cite{zhou2026articraft,chen2026agenticreal2sim,lin2026procedura,lu2025ll3m} can produce a convincing prototype from text or observed objects, but when asked for many assets to cover real-world diversity, they drift into implausible geometries that harm policy performance (Sec.~\ref{sec:exp_asset} and App.~\ref{app:distributions}). Second, demonstrations must be generated across these geometric variations without adding human effort.
Reinforcement learning and planning require task-specific reward design or task models~\cite{ma2023eureka,doi:10.1177/0278364913484072,yin2026omnireset,zhang2026sgs}, and trajectory-augmentation methods need new demonstrations or manually revised task logic for different geometries~\cite{mandlekar2023mimicgen,jiang2024dexmimicen,garrett2024skillmimicgen,xue2025demogen,lin2025cpgen}. Consequently, scaling either assets or demonstrations still requires considerable task-specific engineering.

\begin{figure*}[!t]
    \centering
    \begin{minipage}[b]{0.5084\textwidth}
        \centering
        \includegraphics[width=\linewidth]{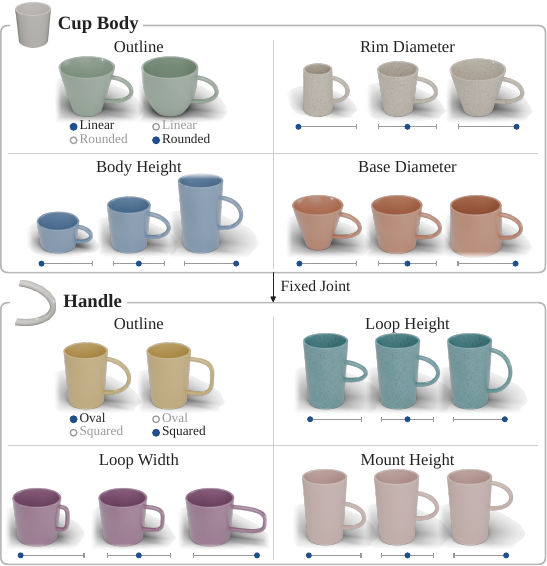}\\[-1mm]
        {\footnotesize (a) Major parameters of the \assetgen mug generator}
    \end{minipage}\hfill
    \begin{minipage}[b]{0.4777\textwidth}
        \centering
        \includegraphics[width=\linewidth]{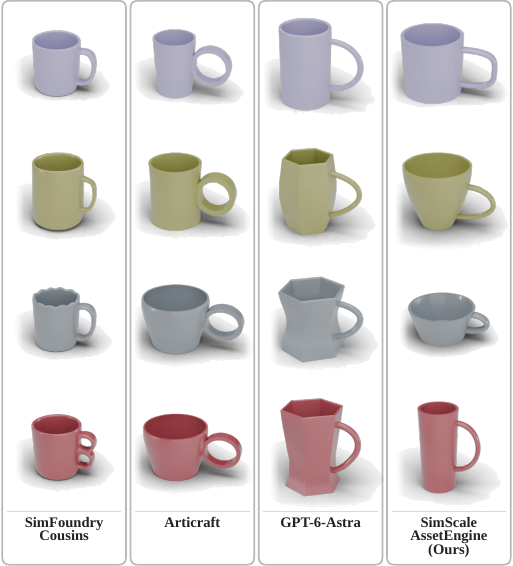}\\[-1mm]
        {\footnotesize (b) Qualitative comparison of different asset generation methods}
    \end{minipage}
    \caption{\textbf{Qualitative comparison: our \system \assetgen generates assets that are both diverse and physically plausible.} (a) Our category-level procedural generator approximates real-world object variations with interpretable parameters for each key feature, grounded by internet images. This representation provides a mechanism to balance geometric diversity and plausibility. (b) Other methods lack this mechanism, so they sacrifice either plausibility or diversity. Physically implausible geometry appears in the body shapes of GPT-6-Astra (bottom three) and in the rim (third) and handle (last) of SimFoundry cousins, while the handles of Articraft are highly similar.}
    \label{fig:asset-qualitative}
    \vspace{-3.7mm}
\end{figure*}

To tackle these challenges, we introduce \system (Fig.~\ref{fig:pull}), an agentic framework that defines category-level procedural generators to produce diverse and physically plausible assets, and generates demonstrations across their geometric variations from one human demonstration.
Our key insight is that a category-level procedural generator is a better representation than individually generated assets for scaling simulation data.
Its explicit parameters, spanning continuous variables for dimensions, proportions, and curved profiles, and discrete variables for counts and design choices, offer a controllable and natural way to expand object diversity.
Furthermore, two critic sub-agents audit each generator to respect the variations observed in internet images of real-world objects.
Diversity therefore grows without sacrificing physical plausibility, whereas generating each asset independently offers no principled mechanism to balance the two.
\system searches web images of the category, encodes the observed variations in a generator, and refines it through agent-driven coverage and geometry reviews.
Given one human demonstration in simulation, \system extracts a storyboard of keyframes and adapts the demonstration across the diverse geometric variations with simulation feedback, scaling it into thousands of trajectories across geometries and poses.
This adaptation process is made easier due to the shared object parametrization that provides ground-truth correspondence between geometric features across assets.

With \system, policies that generalize across object geometries can be trained \textit{entirely} in simulation, with \textit{only one} human demonstration per task. Our contributions are:
\begin{itemize}
    \item We present \system, an agentic framework that automatically generates diverse and physically plausible assets, and generates demonstrations across the wide geometries from one human demonstration per task, enabling zero-shot sim-to-real transfer of RGB visuomotor policies to diverse real-world objects.
    \item We evaluate \system in more than 5,000 real-robot trials on three bimanual tasks and two policy architectures. With one human demonstration per task, \system outperforms policies trained on 60 real-world demonstrations per task by 30.0\% and the strongest simulation baseline with 120 human demonstrations per task by 31.7\%. Scaling \system data steadily improves real-world policy performance without any real-world data, from 57.6\% on average with 10 assets and 1,000 trajectories per task to more than 96\% in all three tasks with 160 assets and 16,000 trajectories per task, totaling over 300 hours of synthetically generated data.
    \item We analyze each component of \system. Compared with assets from existing asset libraries, real-to-sim methods, and agentic asset generation methods, our \system assets significantly improve geometric generalization in simulation and zero-shot sim-to-real transfer. With the same assets, policies trained on our trajectories generated from one human demonstration match those trained on human demonstrations collected for every asset, and outperform policies distilled from task programs written by GPT-6-Astra.
\end{itemize}

\section{Method}
\label{sec:method}

\begin{figure*}[!t]
    \centering
    \includegraphics[width=\textwidth]{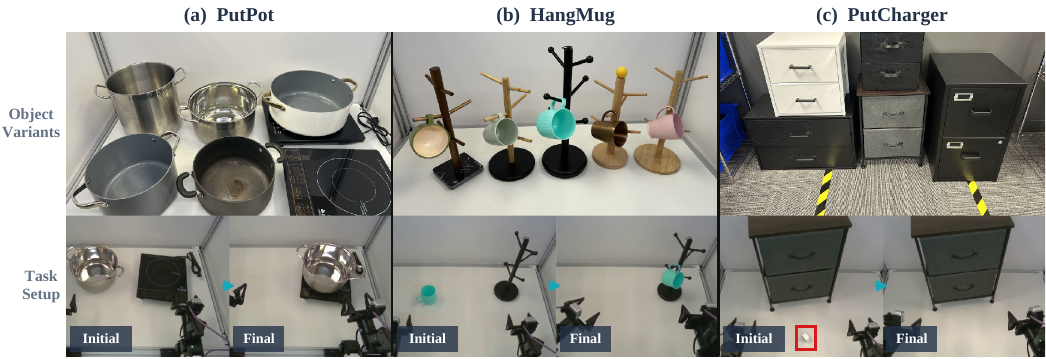}
    \vspace{-12pt}
    \caption{\textbf{Tasks and objects.} \system is evaluated on three bimanual manipulation tasks in the real world, each with five sets of distinct objects.}
    \label{fig:real-world-tasks}
    \vspace{-4mm}
\end{figure*}

\system consists of two coupled engines (Fig.~\ref{fig:agentic-method}). \assetgen authors category-level procedural generators that produce diverse and physically plausible assets, and \demogen adapts one human demonstration across these assets and expands it into thousands of trajectories. In Sec.~\ref{sec:problem-formulation}, we formulate the problem and provide an overview of \system. We then present \assetgen in Sec.~\ref{sec:autoprocgen} and \demogen in Sec.~\ref{sec:autodatagen}. Finally, in Sec.~\ref{sec:policy}, we describe how RGB visuomotor policies are trained on \system-generated data.

\subsection{Problem Setup and Method Overview}
\label{sec:problem-formulation}

\noindent\textbf{Problem Setup.}
Consider a robot manipulation task given by its object categories $\ell$ and optionally a natural-language task description.
Each episode involves objects $O$ with one instance for each task-relevant category, such as a mug and a mug tree.
Real-world objects $O_{\mathrm{real}}$ follow a distribution $p_{\mathrm{real}}(O\mid\ell)$, and our goal is an RGB visuomotor policy $\pi_\psi$ that maximizes the expected real-world success rate,
\begin{equation}
\max_{\psi}\; \mathbb{E}_{O_{\mathrm{real}}\sim p_{\mathrm{real}}(O\mid\ell)}\big[\mathrm{SR}(\pi_\psi, O_{\mathrm{real}})\big],
\label{eq:objective}
\end{equation}
where $\mathrm{SR}(\pi_\psi, O)$ is the success rate on objects $O$ over random initial object poses.
However, $p_{\mathrm{real}}$ is unknown, and real-world demonstrations are both expensive and limited to a narrower, shifted distribution $p_{\mathrm{real,train}}$.
Sim-to-real methods therefore train on simulated objects $O_{\mathrm{sim}}\sim p_{\mathrm{sim}}(O\mid\ell)$ that approximate $p_{\mathrm{real}}$.
A good $p_{\mathrm{sim}}$ covers the geometric variations of $p_{\mathrm{real}}$ while placing little probability on physically implausible objects.

\noindent\textbf{Method Overview.}
\system follows this paradigm and uses no real-world data for training.
It takes $\ell$, an optional task description, and one human demonstration $\tau_{demo}$ teleoperated on one set of objects $O_{\mathrm{sim},0}$ in simulation.
It generates a training set $\mathcal{D}_{\mathrm{train}}$ of successful trajectories over many sampled objects $O_{\mathrm{sim}}$ and initial poses, on which $\pi_\psi$ is trained.
The real-world objects used for evaluation are never used to build $\mathcal{D}_{\mathrm{train}}$.
It consists of two engines (Fig.~\ref{fig:agentic-method}).
\assetgen defines a category-level procedural generator $G:\Theta\rightarrow\mathcal{A}$, where $\Theta$ is a structured parameter space, and each $\theta\in\Theta$ instantiates a task-ready asset $A=G(\theta)$ with its geometric and semantic metadata.
Sampling $\theta\sim p(\theta)$ and applying $G$ defines $p_{\mathrm{sim}}$ for each category.
\demogen takes the sampled assets and the human demonstration, and produces trajectories across the assets.

\subsection{\assetgen: Authoring Procedural Asset Generators}
\label{sec:autoprocgen}

\noindent\textbf{Overview.}
Existing asset generation methods lack a mechanism to balance geometric diversity and physical plausibility when building $p_{\mathrm{sim}}$.
Curated asset libraries~\cite{li2022behaviork,nasiriany2024robocasa,iliash2026artiverse,jin2025artvip} are physically plausible but cover few instances per category.
Real-to-sim methods~\cite{torne2024rialto,dai2024automated,ranawaka2026simfoundry,han2025re3sim,li2026arstag,zhang2026autorsr} build digital twins or cousins of a few real-world objects from $p_{\mathrm{real,train}}$, so their $p_{\mathrm{sim}}$ stays tied to the observed scene.
VLM- or agent-based asset generators~\cite{le2024articulateanything,zhou2026articraft,chen2026agenticreal2sim,lin2026procedura,lu2025ll3m,cao2025physxanything,li2025urdfanything} generate individual assets from text or images.
When prompted for many instances of a category, they drift away from typical prototypes to expand diversity and produce implausible geometries, such as the irregular mug bodies and handles in Fig.~\ref{fig:asset-qualitative}b, so their $p_{\mathrm{sim}}$ places substantial probability on implausible objects.
Our key idea is that a category-level procedural generator significantly improves this balance by building a parameter space that explains the real-world variations observed in internet images.
A new degree of freedom is added only when the current parameterization cannot explain a large portion of the observed objects.
The parameterization is therefore grounded by real-world data, so it expands diversity without pushing the geometry toward implausible instances.
\assetgen defines $G$ with three sub-agents (Fig.~\ref{fig:agentic-method}a): a \emph{Generator Author} that defines the generator grounded by internet images, a \emph{Coverage Critic} that checks coverage and proposes edits for common real-world features the generator cannot explain, and a \emph{Geometry Validator} that checks physical plausibility.
The generator is revised in a loop until it passes both reviews.
We use the mug of HangMug as a running example (Fig.~\ref{fig:agentic-method}a).

\noindent\textbf{Generator Author.}
Given a category description and an optional reference image, the Generator Author actively searches internet images and dimensions to identify geometric variations, feature dependencies, and ranges of measurements.
For mugs, it finds that body height, base diameter, and the height and mount position of the handle loop vary widely across internet examples.
It then writes a coverage plan and implements the generator as a Python program that builds parts and joints with a geometry primitive library and a mesh operation library (Fig.~\ref{fig:agentic-method}a).
Both libraries are design choices, and we build them on the Articraft SDK~\cite{zhou2026articraft}.
The author expresses variations with three types of variables: (1) \emph{continuous absolute variables} sampled over their ordinary ranges, such as mug body height, (2) \emph{continuous relative variables} defined relative to earlier ones, such as base diameter relative to rim diameter, and (3) \emph{discrete variables} over counts or design choices, such as an oval versus a squared handle (Fig.~\ref{fig:asset-qualitative}a).

\noindent\textbf{Coverage Critic.}
The Coverage Critic checks whether the generator covers the real-world variations of the category.
It independently searches for ordinary instances and compares them with sampled objects and key components.
It proposes revisions if common variations are missing or if rare cases bias the generator toward unrealistic samples, and accepts the generator otherwise.
In Fig.~\ref{fig:agentic-method}a, it finds mugs with a rounded body outline that the draft generator cannot produce, and reports them to the Generator Author for revision.

\noindent\textbf{Geometry Validator.}
The Geometry Validator checks whether sampled assets are valid for physical simulation and the task.
It derives two checklists (Fig.~\ref{fig:agentic-method}a): a mesh-quality checklist from the category and physical common sense, and a task-compatibility checklist if a task description is given.
For each item, it writes a function that checks compiled samples with a mesh probe library, which we build on trimesh~\cite{trimesh}.
For HangMug, the task checklist requires an attached, graspable, and hangable handle, so the validator rejects drafts whose handle is detached or is not a closed loop and thus cannot be hung (Fig.~\ref{fig:agentic-method}a).

\noindent\textbf{Asset Sampling.}
Once the generator passes both reviews, we randomly sample parameters $\theta_i\in\Theta$ to obtain diverse and physically plausible assets $\mathcal{A}=\{G(\theta_i)\}_{i=1}^{N}$ for demonstration generation, such as the sampled mugs in Fig.~\ref{fig:agentic-method}a.
An agent can also fit the parameters to a real-world digital twin and perturb them into controllable digital cousins (App.~\ref{app:cousins}).

\subsection{\demogen: Generating Demos Across Assets}
\label{sec:autodatagen}

\noindent\textbf{Overview.}
Typical demonstration generation methods~\cite{mandlekar2023mimicgen,jiang2024dexmimicen, lin2026humanoidmimicgen} expand demonstrations across object poses, but require source demonstrations on each asset. \demogen supplies these demonstrations by adapting one human demonstration $\tau_{demo}$ to novel asset geometries, then uses a pose-expansion method $\mathcal{M}$ to cover new initial poses.
At a high level, our harness constructs a demonstration storyboard, bootstraps a reusable robot execution script, and specializes it across assets before pose expansion (Fig.~\ref{fig:agentic-method}b).
For the HangMug example, $\tau_{demo}$ is teleoperated on one mug and one mug tree sampled from $\mathcal{A}$ (Fig.~\ref{fig:agentic-method}b).

\noindent\textbf{Storyboard construction.}
\phantomsection
\label{sec:source-demos}
First, we distill the single source human demonstration $\tau_{demo}$ into a storyboard, a sequence of keyframes and their timings.
For the HangMug example, the storyboard might include four stages as seen in Fig.~\ref{fig:agentic-method}b: (1) the left arm grasps the mug and (2) hands it to the right arm, (3) the left arm moves its wrist camera close to observe the mug tree, and (4) the right arm hangs the mug.
The camera-view motion involves no contact and would be missed by gripper events alone.

\noindent\textbf{Source-demo bootstrapping.}
Guided by the storyboard, the agent writes a script using CuRobo motion planning~\cite{sundaralingam2023curobo}, gripper control, and collision checking. We randomly sample three assets and have an agent iterate on a single script until the robot successfully completes the task.
For HangMug, the pick, handover, observe, hang, and retreat sequence stays fixed, while grasp and hanging targets follow each mug handle and mug-tree branch (Fig.~\ref{fig:agentic-method}b).

\noindent\textbf{Parallel asset specialization.}
Given a ``base'' script, we launch agents in parallel to adapt the validated script to the remaining assets using the same simulation feedback, generating three source demonstrations per asset from different initial poses. Only executions that complete the task and match the human's pacing enter the source demonstration dataset $\mathcal{D}_{\mathrm{src}}$. App.~\ref{app:demogen-details} describes the execution checks and timing tolerances.

\noindent\textbf{Pose expansion.}
\phantomsection
\label{sec:pose-expansion}
Given a handful of source demonstrations per asset $\mathcal{D}_{\mathrm{src}}$ and stage annotations, $\mathcal{M}$ generates trajectories at new object poses sampled from the task-specific pose distribution and keeps only successful simulation executions. We use DexMimicGen~\cite{jiang2024dexmimicen} as $\mathcal{M}$, which can be replaced by other demonstration generation methods~\cite{mandlekar2023mimicgen,garrett2024skillmimicgen,lin2026humanoidmimicgen}. Agentic script generation covers novel asset geometries, while $\mathcal{M}$ covers object poses, together producing thousands of training trajectories $\mathcal{D}_{\mathrm{train}}$ from one human demonstration.

\subsection{Policy Training}
\label{sec:policy}

We train a visuomotor policy $\pi_\psi$ with behavior cloning on $\mathcal{D}_{\mathrm{train}}$, which maps RGB observations and robot state to a chunk of future actions (Appendix~\ref{app:training}).
To bridge the visual sim-to-real gap, we apply domain randomization~\cite{tobin2017domain,peng2018simtoreal,andrychowicz2020dexterous} to lighting, textures, and colors during data generation.

\section{Experiments}
\label{sec:experiments}

We first describe the experiment setup (Sec.~\ref{sec:exp_setup}). We then evaluate the capabilities of the full \system framework, i.e., whether one human demonstration suffices to train policies that transfer zero-shot to diverse real-world objects, and how they scale with more \system data (Sec.~\ref{sec:exp_system}). Next, we analyze \assetgen by evaluating asset quality and how its assets affect policy generalization in simulation and the real world (Sec.~\ref{sec:exp_asset}). Finally, we analyze \demogen by comparing its trajectories with per-asset human demonstrations and programs written by a coding agent (Sec.~\ref{sec:exp_demo}).

\subsection{Experiment Setup}
\label{sec:exp_setup}

\noindent\textbf{Tasks.}
We evaluate policies on three bimanual manipulation tasks (Fig.~\ref{fig:real-world-tasks}). In PutPot, the robot grasps a pot with both arms and places it upright on a cooktop. In HangMug, the robot picks up a mug, hands it over, and hangs it on a mug tree. In PutCharger, the robot picks up a charger, opens a drawer, places the charger inside, and closes the drawer.

\noindent\textbf{Evaluation protocols.}
In \emph{sim-to-real} evaluation (Secs.~\ref{sec:exp_system}--\ref{sec:exp_demo}), policies are deployed zero-shot on five sets of distinct real-world objects per task. We perform 20 rollouts for each object set, and report the average task score across all five object sets per task.
In \emph{sim-to-sim} evaluation (Sec.~\ref{sec:exp_asset}), we evaluate on unseen assets from different sources: randomly sampled assets from \system-authored procedural generators, held-out RoboCasa assets, and digital twins of all five sets of real-world test objects, and report the average task score over 100 rollouts per asset. RoboCasa has no charger assets, so it is not evaluated on PutCharger. Refer to Apps.~\ref{app:scoring} and~\ref{app:sim} for details.

\noindent\textbf{Training data and baselines.}
We compare \system with policies trained on real-world demonstrations and different simulation baselines.
Real-Demos uses 60 real-world demonstrations per task, 20 on each of three object sets, so three of the five test sets appear in its training data.
Each simulation baseline trains policies on assets from one source: a curated library (RoboCasa~\cite{nasiriany2024robocasa}), agentic asset generators (Articraft~\cite{zhou2026articraft} and GPT-6-Astra~\cite{he2026gptastra}), or a real-to-sim method (SimFoundry twins or cousins~\cite{ranawaka2026simfoundry}). SimFoundry reconstructs a digital twin from a real-world video of a test object set, and augments it with 40 digital cousins.
Except for SimFoundry twins, every simulation dataset uses 40 assets with three source demonstrations per asset, which DexMimicGen~\cite{jiang2024dexmimicen} augments to 100 trajectories per asset.
\system automatically generates all source demonstrations from one human demonstration, whereas the baselines collect the three source demonstrations of each asset by teleoperation, 120 per task for 40 assets.

\noindent\textbf{Policies.}
We train two policy backends on each dataset, GR00T N1.6~\cite{nvidia2025gr00t} and DINOv3 + FMT, which combines a DINOv3 visual backbone~\cite{simeoni2025dinov3} with a flow-matching transformer action head~\cite{lipman2023flowmatching,black2024_0}. Both take three RGB views and the joint and gripper state as input, and predict future action chunks. Refer to App.~\ref{app:training} for training details.

\subsection{SimScale Capabilities}
\label{sec:exp_system}

\noindent\textbf{\system policies transfer zero-shot to diverse real-world objects with far less human effort.}
We compare \system with Real-Demos and all simulation baselines. \system generates all training data in simulation and never sees any real-world objects. As shown in Fig.~\ref{fig:real-world-compact} and Fig.~\ref{fig:real-world-groot} in the appendix, \system performs the best in all tasks and policy backends. Using only one demonstration, \system outperforms Real-Demos, which uses 60 real-world demonstrations per task, by 30.0\% on average, and the strongest simulation baseline, SimFoundry cousins + DexMimicGen, which uses 120 human demonstrations per task, by 31.7\%.

\noindent\textbf{\system policies generalize across diverse object instances without overfitting to a subset of objects.}
Fig.~\ref{fig:real-generalization} compares policy performance on objects seen and unseen by Real-Demos and SimFoundry. Seen objects by SimFoundry are the scenes where the input video is taken for reconstruction. Both baselines degrade significantly from their seen to unseen objects.
Without any real-world demonstrations or careful video capturing, \system outperforms both baselines even on their seen objects, and by even larger margins of 46.0\% and 35.5\% on their unseen objects.
We attribute this to the more diverse and physically plausible training data generated by \system. Geometry augmentation is expensive for Real-Demos in the real world, whereas simulation makes it cheap. SimFoundry, in contrast, remains tied to the specific test scenes it reconstructs.

\noindent\textbf{\system policy performance scales with the number of assets and trajectories.}
\system generates assets automatically and requires only one human demonstration per task, so it scales training data without additional human effort. As shown in Fig.~\ref{fig:real-world-scaling}, real-world performance improves steadily with more assets and trajectories in all three tasks. With 160 assets and 16,000 trajectories, \system achieves a task score of more than 96\% in all three tasks without any real-world data. Even with only 10 assets, \system outperforms on average all simulation baselines trained on 40 assets. These baselines use four times more trajectories than \system and re-collect source demonstrations for each asset, requiring 120 times more human demonstrations. This shows that \system data is of much higher quality.

\subsection{SimScale Analysis: AssetEngine}
\label{sec:exp_asset}

\noindent\textbf{\assetgen assets improve geometric generalization in sim-to-sim settings.}
Fig.~\ref{fig:sim-assets-all} shows policies trained on assets from different sources and evaluated on unseen assets from different sources in simulation, without visual or physical domain gaps. To isolate the effect of assets, all policies, including those trained on \assetgen assets, use per-asset human source demonstrations instead of \demogen. Policies trained on \assetgen assets achieve the highest task score for every task, policy backend, and test asset source, with per-task results in Figs.~\ref{fig:sim-test-assetengine}--\ref{fig:sim-test-twins} in the appendix. Notably, they even outperform policies trained on RoboCasa assets when tested on held-out RoboCasa assets. This illustrates that diverse and physically plausible \assetgen assets boost policy generalization to objects from different sources.

\noindent\textbf{\assetgen assets improve zero-shot sim-to-real transfer.}
Fig.~\ref{fig:real-world-assets} shows policies trained on assets from different sources and evaluated zero-shot on real-world objects, including the first 40 valid assets out of 50 generated by GPT-6-Astra (App.~\ref{app:prompts}), with GR00T N1.6 results in Fig.~\ref{fig:real-world-assets-groot}. Policies trained on \assetgen assets outperform those trained on other asset sources, transferring zero-shot to real-world objects never seen during training. With the same demonstration quality, scaling simulation data on assets sampled from our \system category-level procedural generators yields policies with stronger zero-shot sim-to-real transfer.

\noindent\textbf{\assetgen assets better balance geometric diversity and physical plausibility.}
We evaluate asset quality with a qualitative comparison and a user study. Fig.~\ref{fig:asset-qualitative} shows that \assetgen produces geometrically realistic mugs that vary widely in body shape and handle design, while other methods either drift into physically implausible geometries, such as the irregular body shapes of GPT-6-Astra and the wavy rim and double-loop handle of SimFoundry cousins, or remain highly similar, such as the handles of Articraft. App.~\ref{app:assets} shows the generators and comparisons for more categories. These examples show that existing methods struggle to balance geometric diversity and plausibility, while our category-level procedural generator provides a mechanism for agents to vary geometry while keeping it realistic.

We further conduct a user study to evaluate asset quality quantitatively (Fig.~\ref{fig:asset-preference-all-categories}; details in App.~\ref{app:user-study}). Participants rank sets of 40 assets generated by different methods by two criteria: (1) \textit{geometric realism/plausibility}: how believable the assets are as typical real-world objects, and (2) \textit{geometric diversity}: how widely they span real-world variations, where unrealistic objects do not count. Articraft represents agentic asset generators since its policies outperform those of GPT-6-Astra in two of the three tasks.
On task-relevant object categories, participants rank \system the highest in geometric plausibility and similar to Articraft in diversity, while SimFoundry is the worst on both criteria (Fig.~\ref{fig:asset-preference-all-categories}a). Across 20 categories, including more complex objects, \assetgen assets have significantly higher geometric plausibility than Articraft with comparable geometric diversity (Fig.~\ref{fig:asset-preference-all-categories}b).
Existing methods tend to expand diversity by adding variations at the cost of realism. GPT-6-Astra and Articraft assets show higher variation than \assetgen assets in terms of the Vendi score~\cite{friedman2023vendi} (App.~\ref{app:distributions}), yet Articraft is only comparable to \assetgen in geometric diversity in the user study. This gap indicates that much of their extra variation comes from unrealistic objects far from $p_{\mathrm{real}}$. Policies trained on \assetgen assets outperform those trained on GPT-6-Astra and Articraft assets by 41.5\% and 48.3\% on average (Fig.~\ref{fig:real-world-assets}). These results suggest that both diversity and realism matter for sim-to-real transfer. \assetgen balances both, whereas existing methods tend to sacrifice one for the other.

\subsection{SimScale Analysis: DemoEngine}
\label{sec:exp_demo}

\noindent\textbf{\demogen generates high-quality simulation trajectories matching the quality of human-collected demonstrations.}
Fig.~\ref{fig:real-world-demo} shows sim-to-real policy performance when trained on simulation demonstrations generated by different synthetic trajectory generation methods, where all assets are generated by our \assetgen. Using only one human demonstration, \demogen matches the performance of re-collecting source demonstrations for each new asset.
This illustrates that \demogen preserves the quality of human demonstrations by adapting one human demonstration across diverse object geometries, thus scaling training data across diverse geometry variations without scaling human effort.

\noindent\textbf{\system \demogen generates trajectories that are more suitable for training visuomotor policies than datasets distilled from GPT-6-Astra programs.}
Fig.~\ref{fig:real-world-demo} also compares with GPT-6-Astra Programs, where GPT-6-Astra receives the same human demonstration and iterates in simulation to write a task program for each asset. We then roll out these programs to collect the same number of trajectories for policy training. Although these programs solve the task in simulation with more than 95\% success rate, their policies almost entirely fail in the real world. We observe that these programs produce motions unlike natural human motions, such as unwanted pauses and inconsistent speeds, and often approach and manipulate objects while the wrist camera cannot see the target object. In contrast, \demogen adapts the single human demonstration across objects, preserving natural human motions and intents, resulting in data more suitable for visuomotor policy training.

\section{Related Work}
\label{sec:related_work}

\noindent\textbf{Simulation asset and scene generation.}
Simulation assets are commonly sourced from curated libraries~\cite{li2022behaviork,nasiriany2024robocasa,chen2025robotwin2,iliash2026artiverse,jin2025artvip}, which are physically plausible but offer only a finite set of instances.
Real-to-sim methods reconstruct observed objects and generate constrained variants, sometimes as digital cousins~\cite{torne2024rialto,chen2024urdformer,xia2025drawer,zhao2025rola,dai2024automated,ranawaka2026simfoundry,han2025re3sim}, and recent agentic systems build simulation scenes and data from one image, video, or scan~\cite{li2026arstag,zhang2026autorsr,strong2026real2sem2real}. However, these methods remain tied to specific scenes and miss the rest of the category.
Generative and agentic systems create assets from images or language~\cite{zhou2026articraft,chen2026agenticreal2sim,cao2025physxanything,li2025urdfanything,wang2025kinematify}, and coding agents write programs that model 3D objects~\cite{sun2023threedgpt,lu2025ll3m,lin2026procedura}. Yet they produce one asset per prompt, drift into implausible geometries when scaled to cover real-world diversity, and often produce floating or disconnected parts~\cite{gao2026threedcodebench}.
Procedural generation instead defines a category as a parameterized program, whose parameters systematically vary object-level and part-level geometry~\cite{raistrick2023infinigen,infinigen2024indoors,joshi2025infinigensim,raistrick2026procfunc,lian2025infinitemobility}. Several works~\cite{xue2026doorman,kedia2026simtoolreal} hand-curate simple procedural generators in simulation. However, building procedural generators that cover diverse real-world variations requires substantial graphics and simulation expertise.
In contrast, \system automatically defines a category-level procedural generator grounded in internet examples, and reviews it for coverage and geometric validity, so that sampled assets cover the category's real-world variations while remaining physically plausible.

\noindent\textbf{Synthetic demonstration generation.}
One line of work synthesizes robot demonstrations in simulators using LLM and VLM systems by generating task-specific code~\cite{hua2024gensim2,chen2025robotwin2,hu2026simex}.
Such methods are general in theory, but their behavior can struggle to transfer to the real world.
A complementary approach transforms existing demonstrations:
MimicGen adapts object-centric trajectories to new poses~\cite{mandlekar2023mimicgen, garrett2024skillmimicgen, lin2026humanoidmimicgen}, and DexMimicGen extends this paradigm to bimanual and dexterous manipulation~\cite{jiang2024dexmimicen}.
However, this approach requires new human demonstrations for each new object instance, which makes scaling asset diversity expensive. Recent methods expand one demonstration across object configurations or instances with learned editors, scene edits, reinforcement learning, or reconstructed digital cousins~\cite{qian2026onedemo,jiang2026taskedit,heppert2026real2gen,sharma2026demomimic,strong2026real2sem2real,lum2025human2sim2robot}.
CP-Gen and AffordGen address geometric variation through keypoint--trajectory constraints and affordance correspondences, respectively~\cite{lin2025cpgen,zhang2026affordgen}, but rely on heuristics and were only shown on single-arm tasks.
\system combines the two lines (Sec.~\ref{sec:autodatagen}). Guided by a storyboard of one human demonstration, an agent generates source demonstrations per asset, and DexMimicGen~\cite{jiang2024dexmimicen} expands them across object poses. Agent-generated demonstrations replace the per-asset human demonstrations of MimicGen-style methods and stay closer to human behavior than task-solving code.

\noindent\textbf{Sim-to-real policy learning.}
Domain randomization of visual and physical parameters~\cite{tobin2017domain,peng2018simtoreal,andrychowicz2020dexterous} and training over many object shapes~\cite{chen2023visualdexterity} allow policies trained in simulation to transfer to the real world.
Recent work scales this recipe with large-scale reinforcement learning over diverse resets, distilled into RGB policies~\cite{yin2026omnireset,zhang2026sgs,singh2024dextrahrgb,he2025viral}, and with large synthetic datasets for imitation learning and VLA pretraining~\cite{deshpande2026molmobot,tian2025interndata,deng2025graspvla,ao2026smart}.
More diverse generated scenes improve zero-shot transfer~\cite{choi2026generative3dworlds}, and co-training with real demonstrations narrows the remaining gap~\cite{maddukuri2025cotraining,lei2026mechanisticanalysissimandrealcotraining}.
However, most of these methods scale resets, scenes, rendering, or data volume over a fixed or per-instance set of objects, and behavior cloning on synthetic demonstrations can plateau with more data~\cite{rayyan2026fetchman}.
\system instead scales object geometry with category-level procedural generators, and its policies transfer zero-shot without real-world data.

\section{Conclusion}
\label{sec:conclusion}

We presented \system, an agentic framework that scales up simulation data for sim-to-real robot learning from one human demonstration. Its \assetgen authors category-level procedural generators that balance geometric diversity and physical plausibility, and its \demogen adapts the demonstration across the resulting geometries.
Across more than 5,000 real-robot trials, policies trained only on \system data transfer zero-shot to diverse real-world objects and outperform the strongest baseline by approximately 30\%. Scaling \system data from 10 to 160 assets and from 1,000 to 16,000 trajectories per task steadily improves real-world policy performance to more than 96\% in all three tasks.

\clearpage
\section{Acknowledgements}

We thank Osiriz Durana for help with data collection in the early stages of this project. We also thank Zi-Ang Cao, Congyue Deng, Beining Han, Yifan Hou, Yu Lei, Huihan Liu, Guanzhi Wang, and Josiah Wong for valuable discussions and feedback.

\bibliographystyle{IEEEtran}
\bibliography{references}

\clearpage
\section*{Appendix}
\setcounter{section}{0}
\renewcommand{\thesection}{\Alph{section}}
\renewcommand{\thesubsection}{\thesection.\arabic{subsection}}
\renewcommand{\theHsection}{app.\Alph{section}}
\renewcommand{\theHsubsection}{\theHsection.\arabic{subsection}}
\makeatletter\@ifundefined{thesubsectiondis}{}{\renewcommand{\thesubsectiondis}{\thesection.\arabic{subsection}}}\makeatother

\PinFloat{\the\numexpr\value{page}+3\relax}{1}{
\begin{figure}[t]
    \centering
    \includegraphics[width=\linewidth]{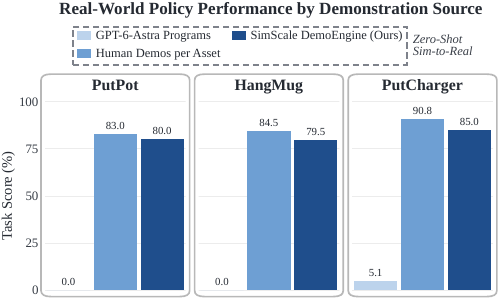}
    \vspace{-6mm}
    \caption{\textbf{Our \system \demogen generates data from one human demonstration on par with human demonstrations re-collected for every asset.}
    Real-world policy performance by synthetic demonstration generation method with GR00T N1.6 on the same 40 \assetgen assets, with DINOv3 + FMT results in Fig.~\ref{fig:real-world-demo}.}
    \label{fig:real-world-demo-groot}
\vspace{-4mm}
\end{figure}
}

\PinFloat{\the\numexpr\value{page}+3\relax}{2}{
\begin{table}[t]
{\centering\footnotesize
\setlength{\tabcolsep}{3pt}
\begin{tabular}{lccc}
\toprule
Category & \shortstack{Continuous\\Error (\%)} & \shortstack{Discrete\\Success (\%)} & IoU (\%) \\
\midrule
Pot & 0.16 & 96.25 & 98.30 \\
Mug & 0.01 & 100.00 & 99.98 \\
Cooktop & 0.01 & 100.00 & 100.00 \\
Mug Tree & $<0.01$ & 100.00 & 100.00 \\
Charger & $<0.01$ & 100.00 & 100.00 \\
\bottomrule
\end{tabular}

\par}
\vspace{3pt}
{\footnotesize\floatcounter\label{tab:recovery}TABLE~\thetable: \textbf{\system accurately recovers the parameters of generator samples.} We evaluate 40 generator samples per category. The table reports mean continuous-parameter error relative to the parameter range, discrete-parameter exact-match rate, and volumetric IoU.\par}
\vspace{-4mm}
\end{table}
}

The appendix gives the details of \system and experiments. In App.~\ref{app:method}, we describe the method details of \system, covering \assetgen, \demogen, additional sim-to-real techniques, policy training, and digital cousin generation. In App.~\ref{app:experiments}, we provide experiment details and additional results for Sec.~\ref{sec:exp_setup}--\ref{sec:exp_demo}.

\section{Method Details}
\label{app:method}

\begin{figure*}[!t]
{\centering
\includegraphics[width=\textwidth]{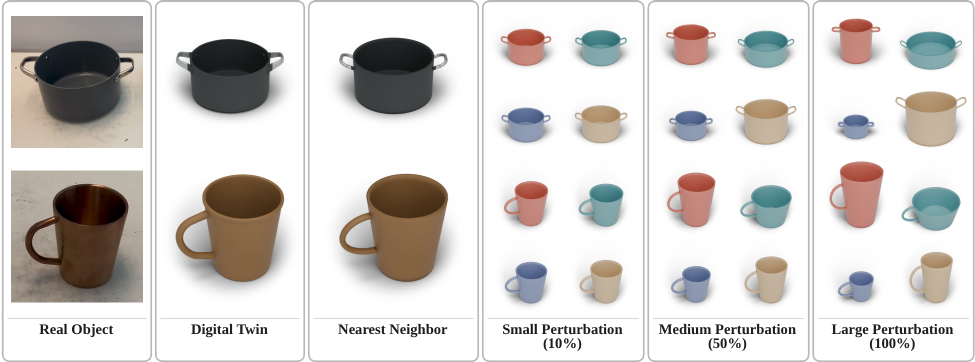}
\par}
\vspace{3pt}
{\footnotesize\floatcounter\label{fig:cousin-sweep}Fig.~\thefigure: \textbf{\system enables controllable digital cousin generation through parameter-space search and perturbation.} From left to right: a real object, its digital twin, the recovered nearest instance in \system procedural generator space, and four cousins at 10\%, 50\%, and 100\% perturbation strength.\par}
\vspace{-4mm}
\end{figure*}

In this section, we describe the method details of \system. We first discuss \assetgen in App.~\ref{app:assetgen-details} and \demogen in App.~\ref{app:demogen-details}. We then describe the additional techniques we use to mitigate the sim-to-real gap in App.~\ref{app:sim2real}, and how we train policies on the generated data in App.~\ref{app:training}.

\subsection{\assetgen Details}
\label{app:assetgen-details}
\label{app:libraries}
\label{app:prompting}

\noindent\textbf{Tool libraries.}
The Generator Author builds parts and joints with a geometry primitive library and a mesh operation library, and the Geometry Validator checks sampled assets with a mesh probe library (Fig.~\ref{fig:agentic-method}a). We implement the first two on top of the Articraft SDK~\cite{zhou2026articraft} and the mesh probe library on top of trimesh~\cite{trimesh}. These libraries are a design choice. We use them because they keep the code written by the agents short and less error-prone, and other libraries can be used in their place. Stronger agents may internalize some of them.

\noindent\textbf{Number of Author-Critic rounds.}
In all experiments, we run the Coverage Critic only once. It reviews the first generator, and the Generator Author revises the generator based on this review. With current agents, we found one round sufficient to cover a wide range of real-world geometric variations. Since the running time of \assetgen is dominated by the Author-Critic loop, this choice also keeps the generation time short. Categories with more complex variations may benefit from additional rounds.

\noindent\textbf{Prompting \assetgen.}
\assetgen only needs a short natural-language request for each category, and we list the requests behind all generators in App.~\ref{app:prompts}. Most requests for the 15 additional categories of the user study are one sentence, so a detailed prompt is not required to obtain a working generator. The more a request specifies how a category varies, the more closely the generator follows it. For example, asking for ``pots with different handle sizes and body proportions'' leads \assetgen to expose these variations as parameters with suitable ranges. Without such guidance, the agent infers the variations from internet references. Since agent outputs are not deterministic, however, the covered variations and their ranges can differ between runs, and stating the desired variations in the request reduces this variance.

\begin{table}[!t]
{\centering
\footnotesize
\setlength{\tabcolsep}{4pt}
\begin{tabular*}{\linewidth}{@{\extracolsep{\fill}}ccc@{}}
\toprule
& Factor & Range \\
\midrule
\multirow[c]{4}{1.75cm}{\centering Materials}
 & Texture Rotation & $[0^\circ,\,360^\circ]$ \\
 & Texture Translation & $[0,\,100]$ \\
 & Color Tint (Objects) & $[0.5,\,1.5]$ \\
 & Color Tint (Workstation, Arms) & $[0.9,\,1.1]$ \\
\midrule
\multirow[c]{3}{1.75cm}{\centering Lighting}
 & Dome Light Intensity & $[300,\,1200]$ \\
 & Dome Color Temperature & $[2000,\,8000]$\,K \\
 & HDRI Rotation & $[0^\circ,\,360^\circ]$ \\
\bottomrule
\end{tabular*}
\par}
\vspace{3pt}
{\footnotesize\floatcounter\label{tab:dr-ranges}TABLE~\thetable: \textbf{Appearance factors randomized during data generation.} Ranges of the material and lighting factors. All factors are resampled every 30 control steps.\par}
\vspace{-4mm}
\end{table}
\noindent\textbf{Requirements are easy to inject.}
Because a generator is a program with explicit parameters, a requirement in the request becomes a constraint on its structure or parameter ranges, and every sampled asset satisfies it by construction. For HangMug, we ask for mug trees with exactly three layers of two opposite branches, with adjacent layers rotated by 90 degrees. For PutCharger, we ask for cabinets with exactly two identical, vertically stacked drawers with pull handles. Similarly, we ask for wardrobes with two identical side-by-side doors. Requirements can also be added after the first generator is authored, for instance to widen the range of a dimension and increase the diversity of its assets, or add a handle type, without authoring the generator from scratch. For the 160-asset setting of the scaling experiment (Sec.~\ref{sec:exp_system}), we use this to widen the ranges of a few major parameters by up to 10\%. Since these are continuous absolute variables and dependent parts are defined by continuous relative variables, the wider ranges keep parts attached and introduce no mesh artifacts such as disconnected parts. This is an advantage of a category-level procedural generator over individually generated assets, and we recommend widening the ranges of major parameters in this way when scaling up the number of assets.

\subsection{\demogen Details}
\label{app:demogen-details}

\noindent\textbf{Storyboard and retargeting.}
Given the source human demonstration, we construct a storyboard using an agent that selects keyframes and records the key interactions as annotations. For retargeting, we map contact-critical poses between corresponding semantic frames on the source and target assets, then search a small deterministic set of nearby candidates. Reachability and collision checks filter these candidates before we rank them by scores such as closeness to the human's pose.
Furthermore, because we are given the ground truth keypoint mappings for our procedurally generated assets, the data generation agent may use these ground truth keypoint mappings for retargeting. If a candidate fails, we revise it before retrying. Between key poses, we replan motion along straight lines in object frames or use CuRobo~\cite{sundaralingam2023curobo} when obstacles require it.
Additionally, we add some randomization during source data demonstration, including ``rest'', ``initial'' and ``observation'' pose randomizations.

\noindent\textbf{Parallel source demonstration generation.}
After we leverage our storyboard and human source demonstration to write a script that's successful  on three diverse assets, we launch and give each agent its own script, assigned jobs, and one GPU.
Each agent refines its script independently to produce successful results.
To accept a rollout, we check both physical task success and pacing versus the storyboard.
In HangMug, the success check would include mug hanging on the target branch.
To preserve the human's pacing, we require the robot to reach each milestone within 2\,s of the human's time, finish within 15\% of the human's duration, and maintain motion through pass-through keyframes.
We also retiming whole paths by arc length to follow the human's smoothed speed profile, and enforce gripper motion to match the teleoperation rate.

\noindent\textbf{Pose expansion settings.}
For pose expansion, we provide DexMimicGen~\cite{jiang2024dexmimicen} with each asset's three source demonstrations and per-arm stage annotations.
In HangMug, these annotations identify the left grasp, handover, and hanging motion. DexMimicGen then transforms and stitches the corresponding motion segments for randomly sampled initial object poses.

\begin{figure*}[!t]
    \centering
    \begin{minipage}[t]{\columnwidth}
        \centering
        \includegraphics[width=\linewidth]{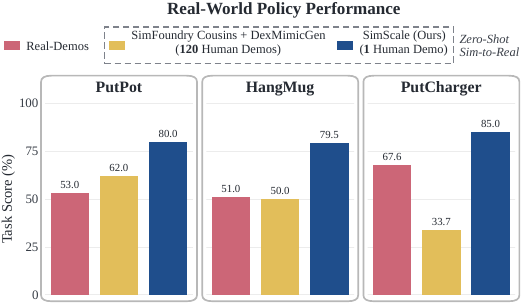}
        \vspace{-6mm}
        \caption{\textbf{\system policies generalize to diverse real-world objects with only one human demonstration and no real-world data.}
        Real-world policy performance with GR00T N1.6, averaged over five object sets per task, with DINOv3 + FMT results in Fig.~\ref{fig:real-world-compact}.}
        \label{fig:real-world-groot}
    \end{minipage}\hfill
    \begin{minipage}[t]{\columnwidth}
        \centering
        \includegraphics[width=\linewidth]{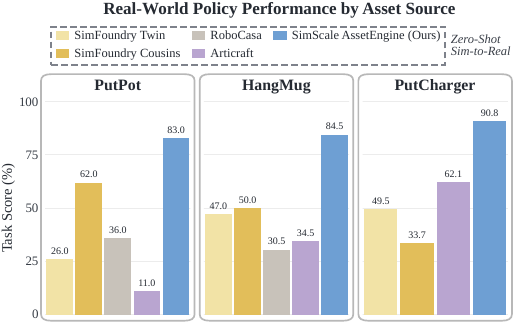}
        \vspace{-6mm}
        \caption{\textbf{Policies trained on assets generated by \system \assetgen show stronger zero-shot transfer to the real world.}
        Real-world policy performance by asset source with GR00T N1.6, with DINOv3 + FMT results in Fig.~\ref{fig:real-world-assets}.}
        \label{fig:real-world-assets-groot}
    \end{minipage}
    \vspace{-4mm}
\end{figure*}

\subsection{Additional Sim-to-Real Techniques}
\label{app:sim2real}

In the main paper, we show that \system mitigates the sim-to-real gap by generating diverse and physically plausible assets and adapting one human demonstration across them. To further mitigate this gap, we randomize appearance during data generation and align the simulated robot with the real one, as described below.

\noindent\textbf{Appearance randomization.}
We randomize appearance while generating demonstrations across object poses using DexMimicGen~\cite{jiang2024dexmimicen}. We assign each asset in simulation a material sampled from a library of materials, with two texture-transform variants per material, and light the scene with an indoor HDRI from Poly Haven~\cite{polyhaven}. We sample all factors uniformly from the ranges in Table~\ref{tab:dr-ranges}. The workstation and the robot arms keep their original materials and receive a small color tint only. We resample all factors every 30 control steps, so each trajectory covers several appearance conditions. We split the material and HDRI libraries into train and evaluation sets. We generate data with the train split, which has 43 materials in 5 categories and 210 HDRIs, and hold out the evaluation split of 17 materials in 7 categories and 10 HDRIs.

\noindent\textbf{Robot and controller alignment.}
We use a bimanual YAM robot with six revolute joints and a parallel-jaw gripper per arm. Each arm runs joint-space PD control with gravity compensation. We align the simulated arms with the real ones in three steps.

First, we correct the mass model. The manufacturer's robot description does not give accurate masses for all links, so we measure the links of our robot and rebuild the gravity-compensation model with the measured values.

Second, we check gravity compensation. The controller adds a gravity-compensation torque to the PD term with a per-joint scale. We use $[1.0, 1.15, 1.2, 1.325, 1.0, 1.0]$ for the re-measured model, higher than the manufacturer's defaults of $[1.0, 1.1, 1.1, 1.2, 1.0, 1.0]$. To check it, we set $k_p = k_d = 0$, which leaves gravity compensation as the only commanded torque, and confirm that each arm holds its pose at ten poses per arm. The compensated arm is effectively gravity-free, so we disable gravity on the simulated arm bodies and keep it on the objects, instead of simulating a compensator.

Third, we identify the controller. We run the real arms with per-joint gains $k_p = [80, 120, 120, 60, 10, 10]$ and $k_d = [5, 5, 5, 1.5, 1.5, 1.5]$, and keep the gripper at the manufacturer's default. We excite all six joints of one arm with a frequency-swept sinusoid of amplitude $0.15$\,rad around $[0^\circ, 60^\circ, 60^\circ, 0^\circ, 0^\circ, 0^\circ]$, sweeping $0.3$ to $2$\,Hz over $12$\,s at $100$\,Hz, and log the commanded and measured joint positions and velocities. We then fit the per-joint stiffness and damping of the simulated PD controller to this response with Levenberg--Marquardt. For each joint, we mean-center the simulated and the recorded position and velocity signals, take their discrete Fourier transforms, and divide the complex difference by the norm of the recorded spectrum, so that joints with different motion amplitudes contribute comparably to the fit. We run this identification once and use the same gains for both arms, ignoring differences between them.

\subsection{Policy Training Details}
\label{app:training}

We train the two policy backends of Sec.~\ref{sec:exp_setup} and Sec.~\ref{sec:policy} with the same inputs and outputs. Each takes three RGB views and the 14-dimensional joint and gripper state as input, and predicts 50-step action chunks of relative arm joint targets and absolute gripper targets. We fine-tune GR00T N1.6 for 40k steps and DINOv3 + FMT for 50k steps, both with a global batch size of 256 and AdamW. For the 160-asset policies of the scaling experiment, we train DINOv3 + FMT for 80k steps. For the Real-Demos baseline, we fine-tune GR00T N1.6 for only 20k steps with a global batch size of 32 to prevent overfitting.

\subsection{Application: Controllable Digital Cousin Generation}
\label{app:cousins}

We present controllable digital cousin generation as a potential downstream application of \system. Real-to-sim methods reconstruct a digital twin of a real-world object and augment it with digital cousins~\cite{dai2024automated,ranawaka2026simfoundry}. Existing methods generate each digital cousin independently with VLMs or agents, without an explicit shape prior for the category. Their digital cousins therefore either drift into physically implausible shapes, such as the wavy rim and double-loop handle of SimFoundry cousins in Fig.~\ref{fig:asset-qualitative}, or stay close to the reconstructed object and cover little of the category. A \system category-level procedural generator provides such a prior, since it varies an object only through named geometric parameters grounded by internet examples. We generate digital cousins in the parameter space of the generator in two steps: (1) we project the digital twin into the parameter space by finding its nearest neighbor in that space, and (2) we perturb the recovered parameters to sample digital cousins. \demogen then adapts one human demonstration to the resulting digital cousins.

\noindent\textbf{Projecting a digital twin into the generator.}
Given an asset and a frozen \system procedural generator, an agent-guided search finds the parameter vector whose generated geometry best matches the observed asset. A multimodal agent configures the geometric search, reviews candidate fits, and revises the search based on geometric errors and visual mismatches. First, the agent inspects the input geometry and generator parameter definitions. It establishes scale and orientation constraints and identifies plausible values for discrete parameters, such as a mug's handle type. The search considers alternative values for uncertain discrete parameters. The agent invokes a fixed search program shared across categories. The program starts from sampled parameter assignments or previous fits, retaining promising candidates for different discrete parameter combinations. For each candidate, it fixes the discrete parameters and refines the continuous parameters. For generator samples, it minimizes differences in the positions and shapes of matched parts using nonlinear least squares. For reconstructed twins, it minimizes surface distance and overall size error using Powell's method. The program reports the best score for each tested discrete parameter combination and renders the leading candidates for agent review. The agent then inspects rendered comparisons and fitting diagnostics for mismatched outlines, misplaced components, or incorrect alignment. The agent revises the search guidance and repeats the search when these checks expose a mismatch. If an observed dimension exceeds an authored range, the agent can expand that parameter's search range without modifying the generator. For example, a mug that remains too short at the upper height bound can trigger a wider height search. The agent guides the search, while the program determines parameter values. The workflow returns the best feasible parameter assignment and generated asset for cousin sampling. See App.~\ref{app:cousin-results} for recovery performance on generator samples and projection results on real-world digital twins.

\noindent\textbf{Sampling digital cousins.}
We sample digital cousins by perturbing the fitted parameters. Both continuous and discrete parameters can vary, and derived parameters are recomputed. Fig.~\ref{fig:cousin-sweep} varies body diameter and height, pot-handle size and tilt, and mug-handle heights along four fixed directions at strengths 0.1, 0.5, and 1.0. Other free parameters remain fixed. Increasing the strength moves each cousin farther along the same direction, giving explicit control over geometric variation.

\section{Additional Experiment Results and Details}
\label{app:experiments}

In this section, we provide additional details and results for the experiments in Sec.~\ref{sec:exp_setup}--\ref{sec:exp_demo}. We first describe how we score real-world trials (App.~\ref{app:scoring}) and report additional real-world and sim-to-sim results (App.~\ref{app:groot}--\ref{app:sim}). We then list the prompts given to each asset generation method (App.~\ref{app:prompts}), evaluate how accurately the nearest-neighbor search of digital cousin generation recovers generator parameters (App.~\ref{app:cousin-results}), and compare the asset distributions of every method (App.~\ref{app:distributions}). Finally, we show the procedural generators of all categories and compare their assets with those of other methods (App.~\ref{app:assets}), and describe the user study (App.~\ref{app:user-study}).

\subsection{Real-World Scoring\texorpdfstring{ (Sec.~\ref{sec:exp_setup})}{}}
\label{app:scoring}

We compute the task scores of Sec.~\ref{sec:exp_setup} as follows. In PutPot, a trial scores 1 if the pot is placed upright on the cooktop and released, and 0 otherwise. In HangMug, a trial scores 0.5 once the mug is handed over, and 1 once it is hung on the tree and released. In PutCharger, a trial scores 0.3 once the charger is picked up, 0.7 once the drawer is opened, and 1 once the charger is placed inside and the drawer is closed. We run 20 rollouts for each object set and report the average score over all rollouts.

\subsection{Additional Real-World Results\texorpdfstring{ (Sec.~\ref{sec:exp_system}--\ref{sec:exp_demo})}{}}
\label{app:groot}

\noindent\textbf{\system outperforms Real-Demos and SimFoundry cousins.}
Fig.~\ref{fig:real-world-groot} repeats the experiment of Fig.~\ref{fig:real-world-compact} with GR00T N1.6 as the policy backend. Using only one human demonstration and no real-world data, \system outperforms Real-Demos, which uses 60 real-world demonstrations per task, by 24.3\% on average, and SimFoundry cousins + DexMimicGen, which uses 120 human demonstrations per task, by 32.9\%. This is consistent with the results with DINOv3 + FMT in Sec.~\ref{sec:exp_system}.

\noindent\textbf{\assetgen assets outperform all other asset sources.}
Fig.~\ref{fig:real-world-assets-groot} repeats the experiment of Fig.~\ref{fig:real-world-assets} with GR00T N1.6 as the policy backend. Policies trained on \assetgen assets outperform every other asset source in every task, and outperform SimFoundry cousins, the strongest other asset source, by 37.5\% on average. This is consistent with the results with DINOv3 + FMT in Sec.~\ref{sec:exp_asset}.

\noindent\textbf{\demogen performs on par with human demonstrations re-collected for every asset.}
Fig.~\ref{fig:real-world-demo-groot} repeats the experiment of Fig.~\ref{fig:real-world-demo} with GR00T N1.6 as the policy backend. From one human demonstration, \demogen performs within 5\% of re-collecting human demonstrations for every asset on average, while policies trained on GPT-6-Astra programs almost entirely fail. This is consistent with the results with DINOv3 + FMT in Sec.~\ref{sec:exp_demo}.

\makeatletter
\setlength{\@fptop}{0pt}\setlength{\@fpsep}{16pt}\setlength{\@fpbot}{0pt plus 1fil}
\setlength{\@dblfptop}{0pt}\setlength{\@dblfpsep}{16pt}\setlength{\@dblfpbot}{0pt plus 1fil}
\makeatother

\begin{figure*}[!t]
    \centering
    \includegraphics[width=\textwidth]{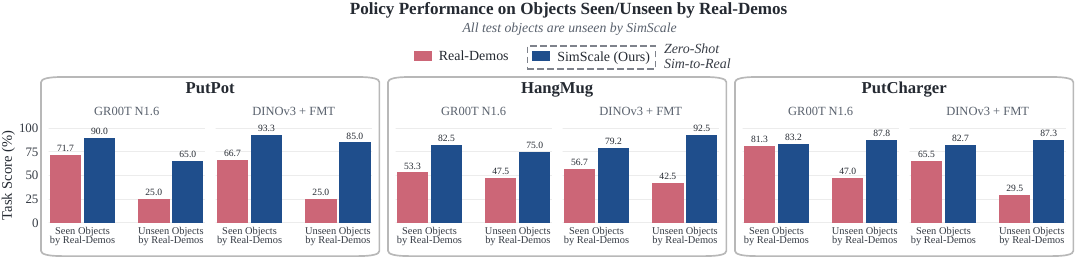}
    \vspace{-6mm}
    \caption{\textbf{\system generalizes to various unseen real-world objects, while baselines degrade sharply on unseen objects.}
    Real-world policy performance on objects seen and unseen by Real-Demos, per task and policy backend, with averages in Fig.~\ref{fig:real-generalization}. Seen objects are the three of the five test objects with Real-Demos demonstrations in each task.}
    \label{fig:seen-unseen-real-demos}

\end{figure*}

\begin{figure*}[t]
    \centering
    \includegraphics[width=\textwidth]{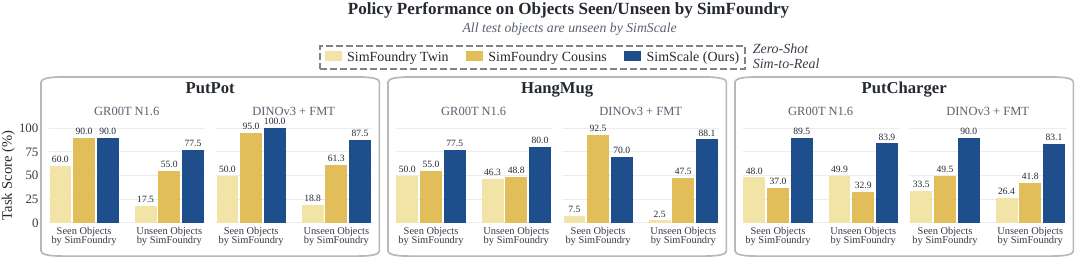}
    \vspace{-6mm}
    \caption{\textbf{\system generalizes to various unseen real-world objects, while baselines degrade sharply on unseen objects.}
    Real-world policy performance on objects seen and unseen by SimFoundry, per task and policy backend, with averages in Fig.~\ref{fig:real-generalization}. The seen object is the test object captured in the input video of SimFoundry in each task.}
    \label{fig:seen-unseen-simfoundry}
\end{figure*}

\noindent\textbf{\system outperforms Real-Demos and SimFoundry on objects unseen by them in every task and policy backend.}
Figs.~\ref{fig:seen-unseen-real-demos} and~\ref{fig:seen-unseen-simfoundry} break down Fig.~\ref{fig:real-generalization} by task and policy backend. Both baselines perform worse on unseen objects than on the objects they observe, while \system outperforms both baselines on unseen objects in every task and policy backend. This is consistent with Sec.~\ref{sec:exp_system}.
Real-Demos collects real-world demonstrations on three of the five test object sets in each task, while SimFoundry takes a real-world video of one test object set in each task as input.
In contrast, \system does not observe any test object.

\makeatletter
\setlength{\@dblfptop}{0pt}\setlength{\@dblfpsep}{16pt}\setlength{\@dblfpbot}{0pt plus 1fil}
\makeatother

\setcounter{dbltopnumber}{1}
\subsection{Detailed Sim-to-Sim Results\texorpdfstring{ (Sec.~\ref{sec:exp_asset})}{}}
\label{app:sim}

\begin{figure*}[tp]
    \centering
    \includegraphics[width=\textwidth]{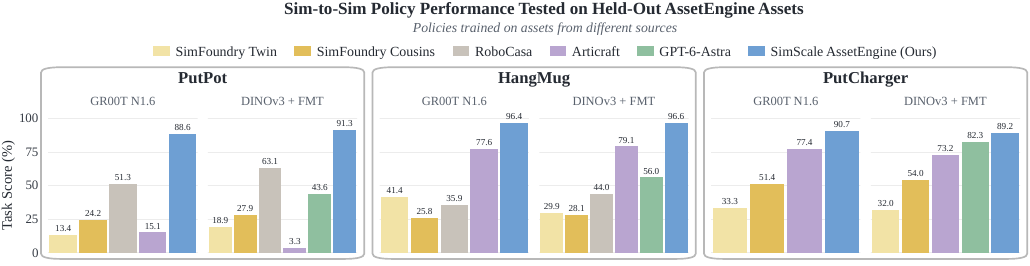}
    \vspace{-6mm}
    \caption{\textbf{Policies trained on assets generated by \system \assetgen exhibit stronger geometric generalization on held-out \assetgen assets.}
    Sim-to-sim performance of policies trained on assets from different sources and tested on held-out \assetgen assets, per task and policy backend, with averages in Fig.~\ref{fig:sim-assets-all}.}
    \label{fig:sim-test-assetengine}

    \vspace{10pt}
    \includegraphics[width=\textwidth]{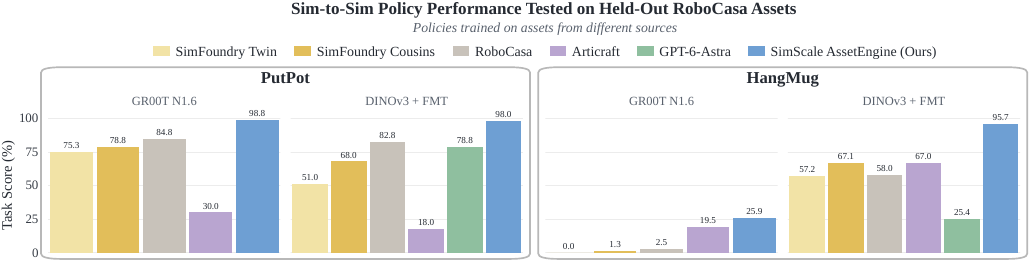}
    \vspace{-6mm}
    \caption{\textbf{Policies trained on assets generated by \system \assetgen exhibit stronger geometric generalization on held-out RoboCasa assets.}
    Sim-to-sim performance of policies trained on assets from different sources and tested on held-out RoboCasa assets, per task and policy backend, with averages in Fig.~\ref{fig:sim-assets-all}.}
    \label{fig:sim-test-robocasa}
\end{figure*}

\begin{figure*}[tp]
    \centering
    \includegraphics[width=\textwidth]{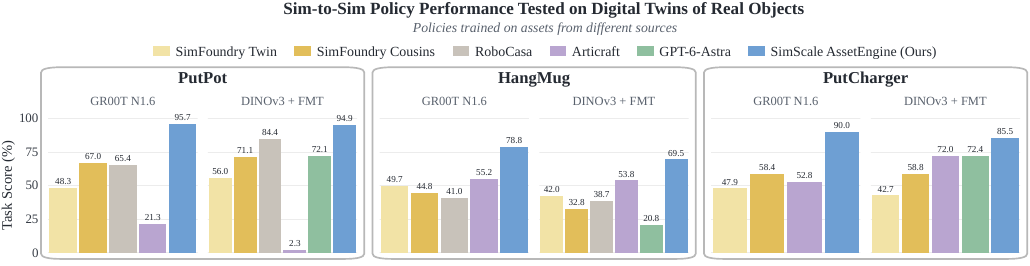}
    \vspace{-6mm}
    \caption{\textbf{Policies trained on assets generated by \system \assetgen exhibit stronger geometric generalization on digital twins of real objects.}
    Sim-to-sim performance of policies trained on assets from different sources and tested on digital twins of real objects, per task and policy backend, with averages in Fig.~\ref{fig:sim-assets-all}.}
    \label{fig:sim-test-twins}
\end{figure*}

\noindent\textbf{Policies trained on \assetgen assets achieve the highest task score in every task, policy backend, and test asset source.}
Figs.~\ref{fig:sim-test-assetengine}--\ref{fig:sim-test-twins} break down Fig.~\ref{fig:sim-assets-all} by task and policy backend, and lead to the same conclusion as Sec.~\ref{sec:exp_asset}. Since RoboCasa has no charger assets, we do not train or test RoboCasa policies on PutCharger. RoboCasa has fewer than 40 assets per category, so we hold out 20\% of the assets of each category as test assets, and rescale the remaining assets to obtain 40 training assets.

\subsection{Asset Generation Setup and Prompts}
\label{app:prompts}

\noindent\textbf{Agent and generation time.}
For all asset comparisons, we run both \assetgen and Articraft with GPT-5.6 at high reasoning effort. Over the 20 categories, \assetgen takes 20--60 minutes to author the procedural generator of one category, while Articraft takes 8--16 minutes to author one asset.

\noindent\textbf{Prompts.}
Tables~\ref{tab:prompts-task} and~\ref{tab:prompts-additional} list the prompts we give each method, for the task-relevant categories and for the 15 additional categories of the user study, respectively. SimFoundry cousins take a real-world video as input instead of a prompt.

\subsection{Accuracy of Nearest-Neighbor Search in \assetgen Generators}
\label{app:cousin-results}

We evaluate the accuracy of the nearest-neighbor search of App.~\ref{app:cousins} on generator samples with known parameters and on our real-world digital twins.

\begin{figure*}[tp]
\begin{minipage}[t]{\columnwidth}
{\centering
\includegraphics[width=\linewidth]{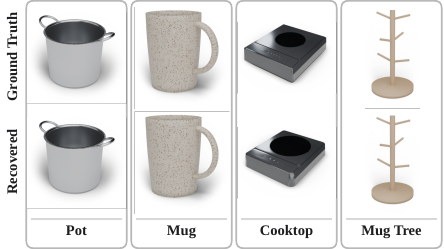}
\par}
\vspace{3pt}
{\footnotesize\floatcounter\label{fig:recovery-overlays}Fig.~\thefigure: \textbf{\system accurately recovers generator samples in the parameter space of \assetgen procedural generators.} Top: ground-truth samples. Bottom: recovered samples.\par}
\end{minipage}\hfill
\begin{minipage}[t]{\columnwidth}
{\centering
\includegraphics[width=\linewidth]{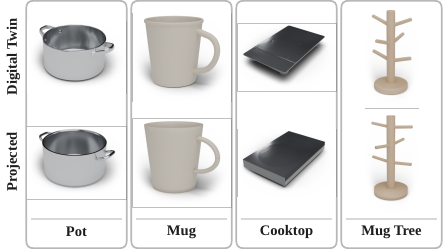}
\par}
\vspace{3pt}
{\footnotesize\floatcounter\label{fig:twin-overlays}Fig.~\thefigure: \textbf{\system projects real-world digital twins into the parameter space of \assetgen procedural generators.} Top: SimFoundry digital twins. Bottom: their projections in the \assetgen procedural generators.\par}
\end{minipage}
\end{figure*}

\noindent\textbf{Recovery of the original generator samples.}
We randomly sample 40 assets from each task-relevant category and run our pipeline to recover the parameters. Cabinets are excluded since a large portion of its parameters are interior parameters. For each continuous parameter, the error is $|\hat{\theta}-\theta|/(b-a)$, where $\theta$ and $\hat{\theta}$ are the ground-truth and recovered values used to build the geometry, and $a$ and $b$ are the generator's sampling bounds for the ground-truth asset. Table~\ref{tab:recovery} shows the performance of our parameter-recovery pipeline, including continuous-parameter recovery error, discrete-parameter recovery success rates, and volumetric IoU. These results show that our pipeline accurately recovers continuous and discrete parameters and closely reproduces the geometry of generator samples. Fig.~\ref{fig:recovery-overlays} shows qualitative recovery examples.

\noindent\textbf{\system accurately projects digital twins of real-world objects into generator parameter space.}
Fig.~\ref{fig:twin-overlays} shows qualitative results of digital twins of real-world objects and their projected assets within \system procedural generator space. The fitted parameters provide starting points for controllable digital cousin generation (Fig.~\ref{fig:cousin-sweep}).

\subsection{More Qualitative Asset Comparisons}
\label{app:assets}

\begin{figure*}[!t]
    \centering
    \begin{minipage}[b]{0.5458\textwidth}
        \centering
        \includegraphics[width=\linewidth]{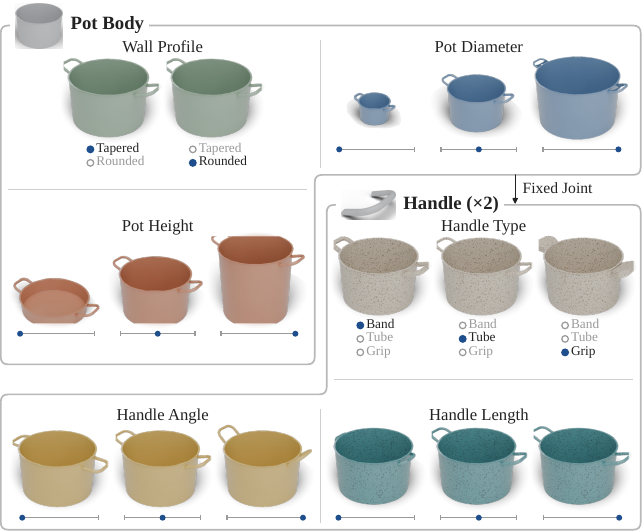}\\[-1mm]
        {\footnotesize (a) Major parameters of the \assetgen cooking pot generator}
    \end{minipage}\hfill
    \begin{minipage}[b]{0.4422\textwidth}
        \centering
        \includegraphics[width=\linewidth]{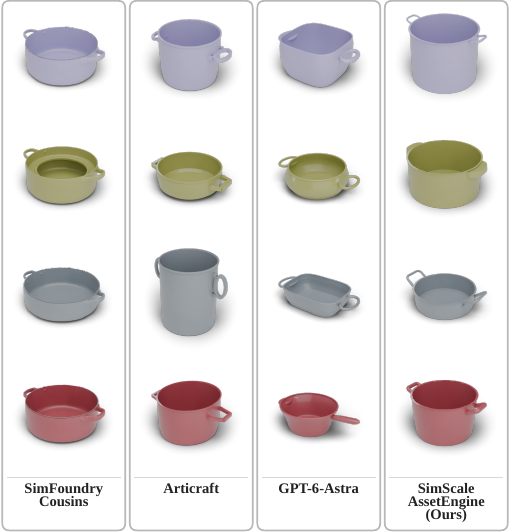}\\[-1mm]
        {\footnotesize (b) Qualitative comparison of different asset generation methods}
    \end{minipage}
    \caption{\textbf{Pot.} (a) The \assetgen generator varies the wall profile, diameter, and height of the pot, and the type, angle, and length of its handles. (b) SimFoundry cousins are shallow and similar, GPT-6-Astra produces a baking dish (third) and a long-handled saucepan (last), and Articraft generates unrealistic handles.}
    \label{fig:assets-pot}
\end{figure*}

\begin{figure*}[!t]
    \centering
    \begin{minipage}[b]{0.4529\textwidth}
        \centering
        \includegraphics[width=\linewidth]{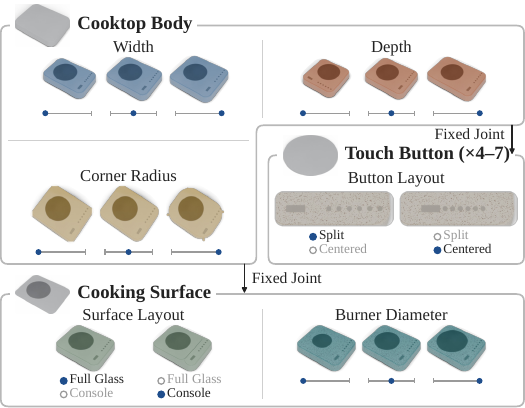}\\[-1mm]
        {\footnotesize (a) Major parameters of the \assetgen cooktop generator}
    \end{minipage}\hfill
    \begin{minipage}[b]{0.5351\textwidth}
        \centering
        \includegraphics[width=\linewidth]{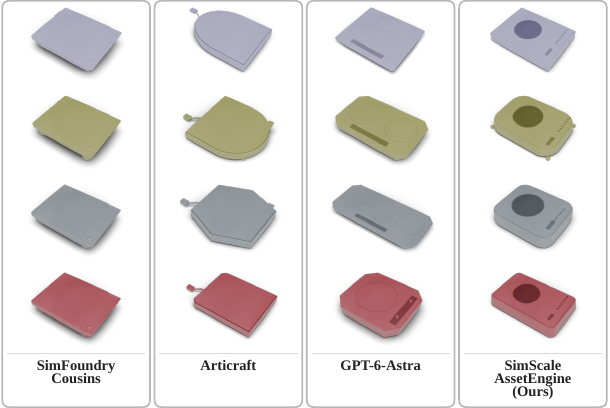}\\[-1mm]
        {\footnotesize (b) Qualitative comparison of different asset generation methods}
    \end{minipage}
    \caption{\textbf{Cooktop.} (a) The \assetgen generator varies the size, corner radius, button layout, surface layout, and burner size of the cooktop. (b) SimFoundry cousins are highly similar, GPT-6-Astra tends to generate rectangular and elongated cooktops, and Articraft produces implausible shapes such as D-shaped and hexagonal tops (first three).}
    \label{fig:assets-cooktop}
\end{figure*}

\begin{figure*}[!t]
    \centering
    \begin{minipage}[b]{0.4892\textwidth}
        \centering
        \includegraphics[width=\linewidth]{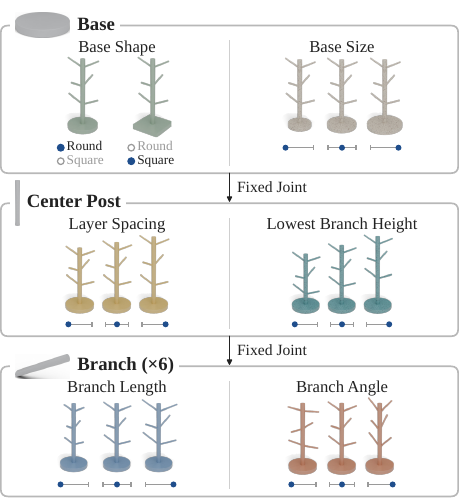}\\[-1mm]
        {\footnotesize (a) Major parameters of the \assetgen mug tree generator}
    \end{minipage}\hfill
    \begin{minipage}[b]{0.4988\textwidth}
        \centering
        \includegraphics[width=\linewidth]{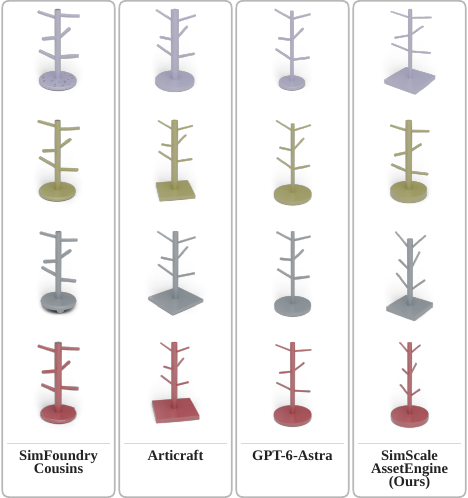}\\[-1mm]
        {\footnotesize (b) Qualitative comparison of different asset generation methods}
    \end{minipage}
    \caption{\textbf{Mug tree.} (a) The \assetgen generator varies the base shape and size, the layer spacing and lowest branch height, and the length and angle of the branches. (b) Mug trees from each method are more similar than in other categories because the prompts of Articraft, GPT-6-Astra, and \assetgen add explicit constraints, three layers of two opposite branches with adjacent layers rotated by 90 degrees (Table~\ref{tab:prompts-task}).}
    \label{fig:assets-mug-tree}
\end{figure*}

\begin{figure*}[!t]
    \centering
    \begin{minipage}[b]{0.6091\textwidth}
        \centering
        \includegraphics[width=\linewidth]{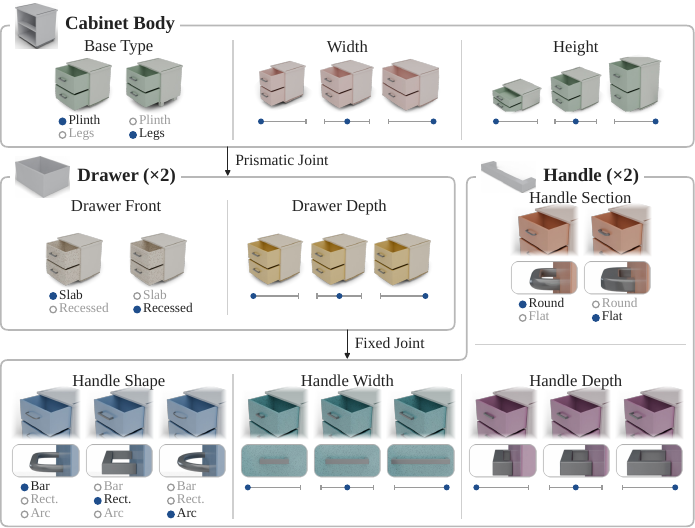}\\[-1mm]
        {\footnotesize (a) Major parameters of the \assetgen two-drawer cabinet generator}
    \end{minipage}\hfill
    \begin{minipage}[b]{0.3789\textwidth}
        \centering
        \includegraphics[width=\linewidth]{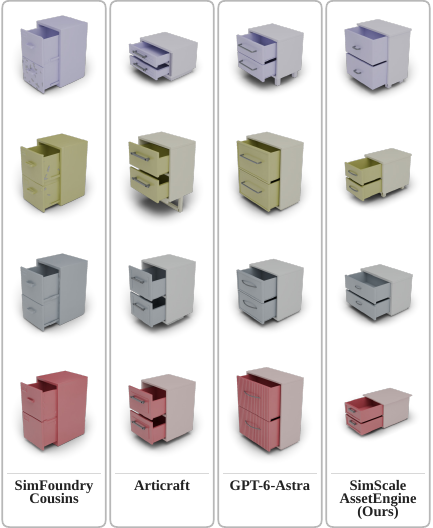}\\[-1mm]
        {\footnotesize (b) Qualitative comparison of different asset generation methods}
    \end{minipage}
    \caption{\textbf{Two-drawer cabinet.} (a) The \assetgen generator varies the base, size, drawer fronts, and drawer depth of the cabinet, and the shape, section, width, and depth of its handles. (b) SimFoundry cousins keep the proportions of the observed cabinet, while Articraft and GPT-6-Astra repeatedly generate very elongated handles that span a large part of the drawer front.}
    \label{fig:assets-drawer-cabinet}
\end{figure*}

\begin{figure*}[!t]
    \centering
    \begin{minipage}[b]{0.5080\textwidth}
        \centering
        \includegraphics[width=\linewidth]{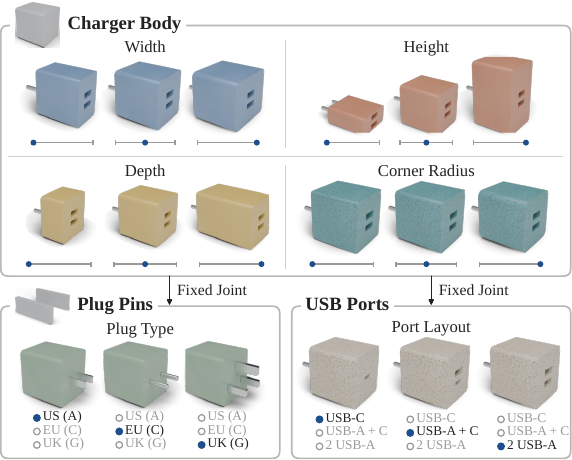}\\[-1mm]
        {\footnotesize (a) Major parameters of the \assetgen phone charger generator}
    \end{minipage}\hfill
    \begin{minipage}[b]{0.4800\textwidth}
        \centering
        \includegraphics[width=\linewidth]{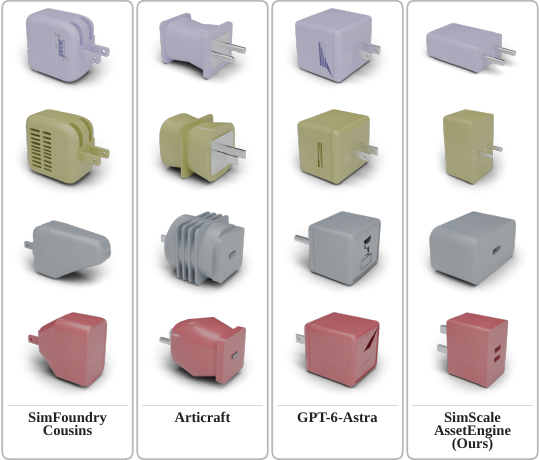}\\[-1mm]
        {\footnotesize (b) Qualitative comparison of different asset generation methods}
    \end{minipage}
    \caption{\textbf{Charger.} (a) The \assetgen generator varies the plug standard, USB ports, and body shape of the charger. (b) SimFoundry cousins warp one body into wedge shapes (last two), Articraft produces a pinched body (first) and radiator fins (third), and GPT-6-Astra bodies are near-identical cubes with torn surfaces.}
    \label{fig:assets-charger}
\end{figure*}

Fig.~\ref{fig:asset-qualitative} shows the \assetgen mug generator and compares mugs from different asset generation methods. Figs.~\ref{fig:assets-pot}--\ref{fig:assets-wardrobe} show the same for the other five task-relevant categories (cooking pots, cooktops, mug trees, two-drawer cabinets, and phone chargers) and the 15 additional categories of the user study (App.~\ref{app:user-study}).
In (a), we show the major parameters. In (b), we show qualitative comparisons of assets generated by different methods.

\subsection{User Study}
\label{app:user-study}
This section describes the details of the user study reported in Fig.~\ref{fig:asset-preference-all-categories} (Sec.~\ref{sec:exp_asset}).

\begin{figure*}[!t]
    \centering
    \begin{minipage}[b]{0.6762\textwidth}
        \centering
        \includegraphics[width=\linewidth]{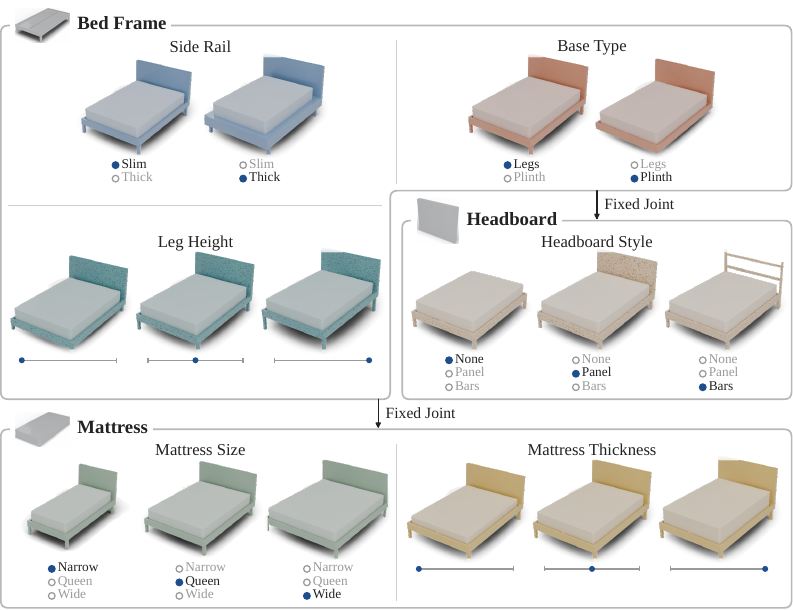}\\[-1mm]
        {\footnotesize (a) Major parameters of the \assetgen bed generator}
    \end{minipage}\hfill
    \begin{minipage}[b]{0.3118\textwidth}
        \centering
        \includegraphics[width=\linewidth]{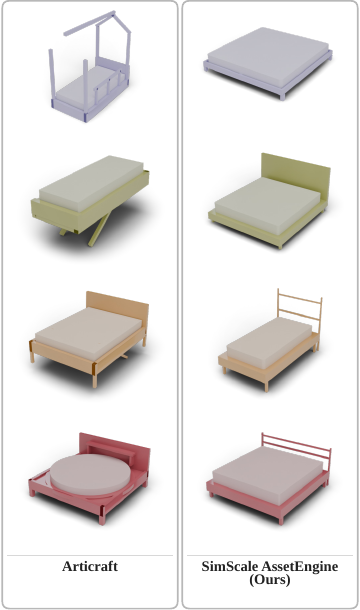}\\[-1mm]
        {\footnotesize (b) Qualitative comparison of different asset generation methods}
    \end{minipage}
    \caption{\textbf{Bed.} (a) The \assetgen generator varies the mattress size and thickness, headboard, side rails, base, and leg height of the bed. (b) Articraft produces a house-shaped frame with a disconnected roof beam (first) and a bed on misaligned legs (second).}
    \label{fig:assets-bed}
\end{figure*}

\begin{figure*}[!t]
    \centering
    \begin{minipage}[b]{0.7118\textwidth}
        \centering
        \includegraphics[width=\linewidth]{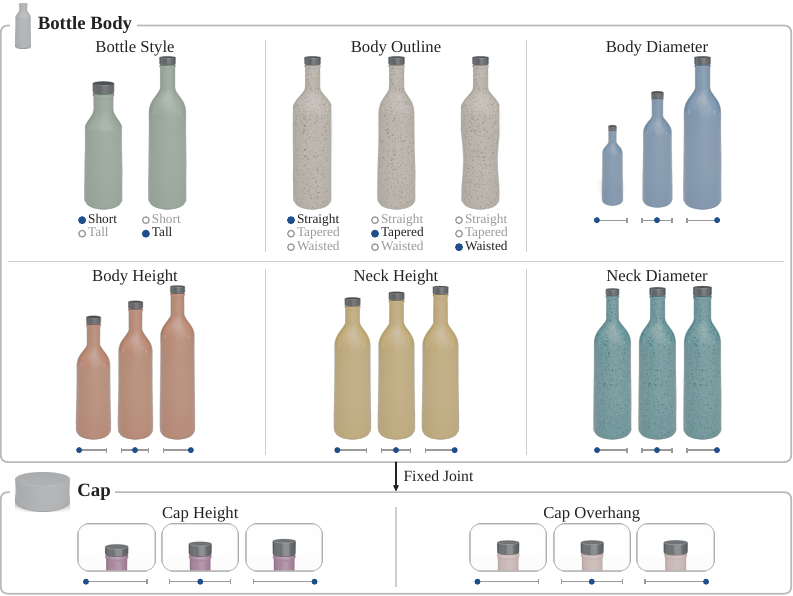}\\[-1mm]
        {\footnotesize (a) Major parameters of the \assetgen bottle generator}
    \end{minipage}\hfill
    \begin{minipage}[b]{0.2762\textwidth}
        \centering
        \includegraphics[width=\linewidth]{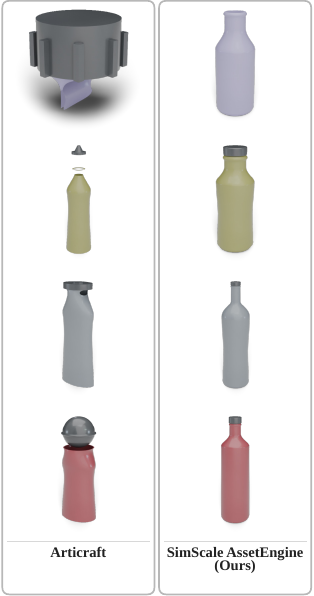}\\[-1mm]
        {\footnotesize (b) Qualitative comparison of different asset generation methods}
    \end{minipage}
    \caption{\textbf{Bottle.} (a) The \assetgen generator varies the style, outline, and proportions of the bottle, and the height and overhang of its cap. (b) The first Articraft sample is not a bottle, the cap and a ring of the second float above its neck, the third has a sheared body, and the last is closed by a sphere.}
    \label{fig:assets-bottle}
\end{figure*}

\begin{figure*}[!t]
    \centering
    \begin{minipage}[b]{0.6784\textwidth}
        \centering
        \includegraphics[width=\linewidth]{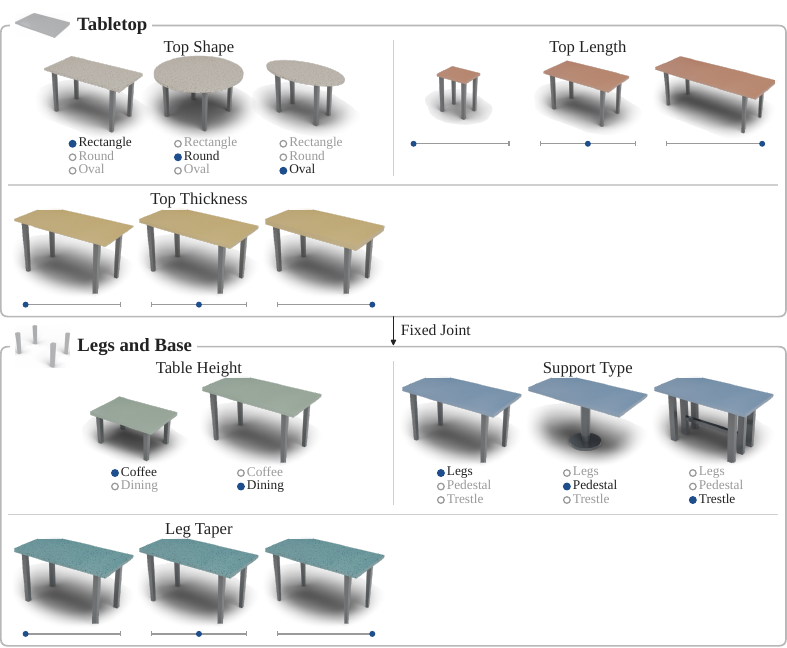}\\[-1mm]
        {\footnotesize (a) Major parameters of the \assetgen table generator}
    \end{minipage}\hfill
    \begin{minipage}[b]{0.3096\textwidth}
        \centering
        \includegraphics[width=\linewidth]{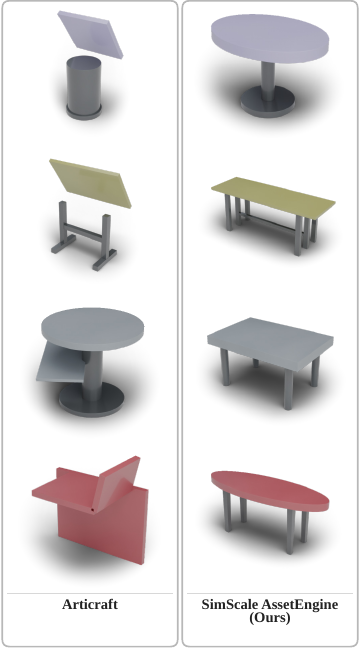}\\[-1mm]
        {\footnotesize (b) Qualitative comparison of different asset generation methods}
    \end{minipage}
    \caption{\textbf{Table.} (a) The \assetgen generator varies the height, top shape and size, support, and leg taper of the table. (b) The tops of the first two Articraft tables float tilted above their bases, the third has a stray panel, and the last is made of intersecting slabs.}
    \label{fig:assets-table}
\end{figure*}

\begin{figure*}[!t]
    \centering
    \begin{minipage}[b]{0.6531\textwidth}
        \centering
        \includegraphics[width=\linewidth]{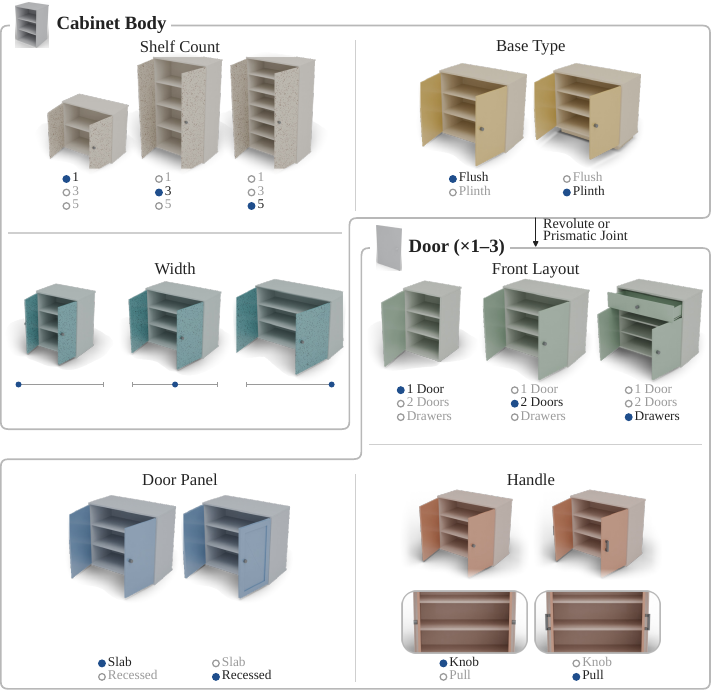}\\[-1mm]
        {\footnotesize (a) Major parameters of the \assetgen cabinet generator}
    \end{minipage}\hfill
    \begin{minipage}[b]{0.3349\textwidth}
        \centering
        \includegraphics[width=\linewidth]{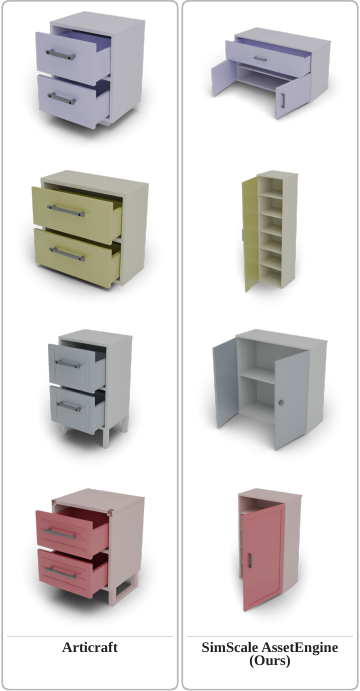}\\[-1mm]
        {\footnotesize (b) Qualitative comparison of different asset generation methods}
    \end{minipage}
    \caption{\textbf{Cabinet.} (a) The \assetgen generator varies the front layout, shelf count, door panel, handles, base, and width of the cabinet. (b) All four Articraft cabinets are two-drawer nightstands.}
    \label{fig:assets-cabinet}
\end{figure*}

\begin{figure*}[!t]
    \centering
    \begin{minipage}[b]{0.7506\textwidth}
        \centering
        \includegraphics[width=\linewidth]{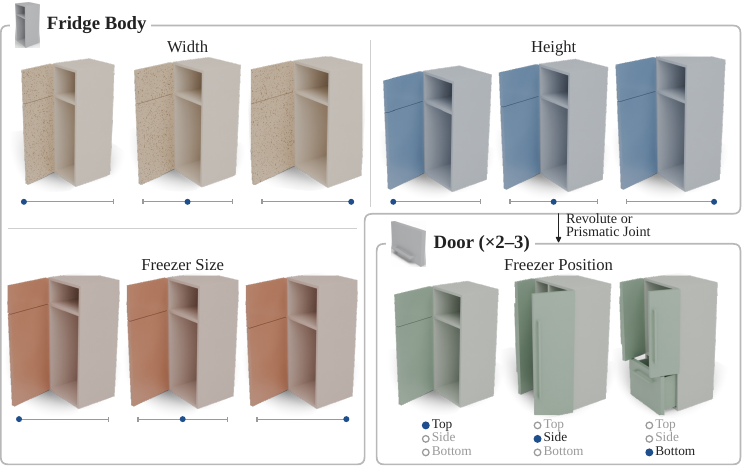}\\[-1mm]
        {\footnotesize (a) Major parameters of the \assetgen refrigerator generator}
    \end{minipage}\hfill
    \begin{minipage}[b]{0.2374\textwidth}
        \centering
        \includegraphics[width=\linewidth]{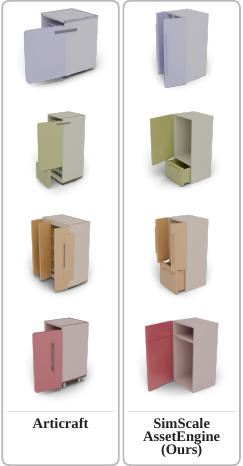}\\[-1mm]
        {\footnotesize (b) Qualitative comparison of different asset generation methods}
    \end{minipage}
    \caption{\textbf{Refrigerator.} (a) The \assetgen generator varies the freezer position and size, width, and height of the refrigerator. (b) The doors of the first two Articraft refrigerators have no handles, and the drawer of the second is detached from its body. The third and fourth place handles near the hinge instead of at the free edge of the door.}
    \label{fig:assets-refrigerator}
\end{figure*}

\begin{figure*}[!t]
    \centering
    \begin{minipage}[b]{0.4786\textwidth}
        \centering
        \includegraphics[width=\linewidth]{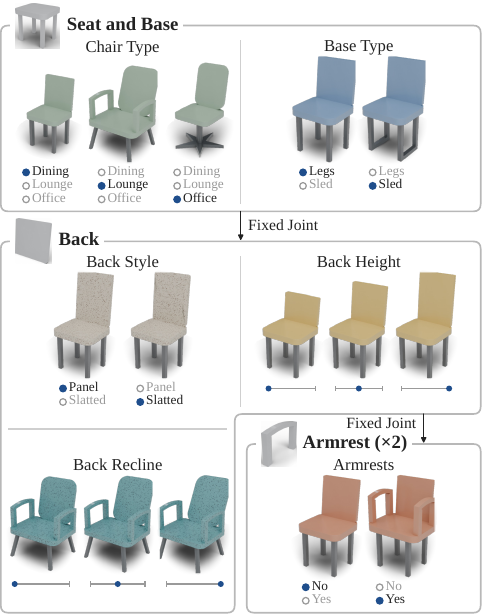}\\[-1mm]
        {\footnotesize (a) Major parameters of the \assetgen chair generator}
    \end{minipage}\hfill
    \begin{minipage}[b]{0.5094\textwidth}
        \centering
        \includegraphics[width=\linewidth]{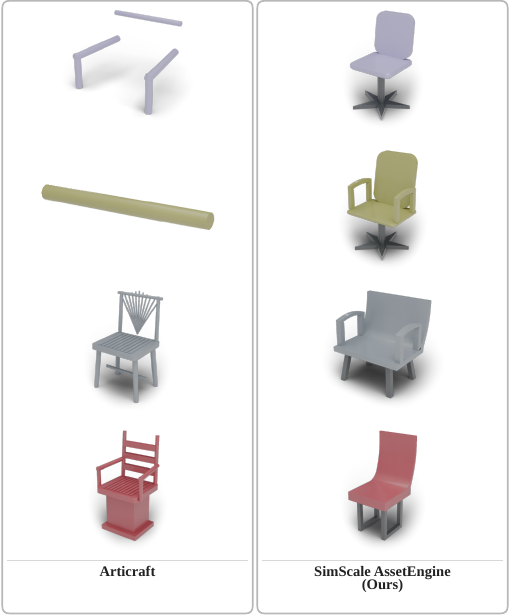}\\[-1mm]
        {\footnotesize (b) Qualitative comparison of different asset generation methods}
    \end{minipage}
    \caption{\textbf{Chair.} (a) The \assetgen generator varies the type, back, base, armrests, and recline of the chair. (b) The first Articraft chair is a set of floating bars, and the second is a single bar.}
    \label{fig:assets-chair}
\end{figure*}

\begin{figure*}[!t]
    \centering
    \begin{minipage}[b]{0.6314\textwidth}
        \centering
        \includegraphics[width=\linewidth]{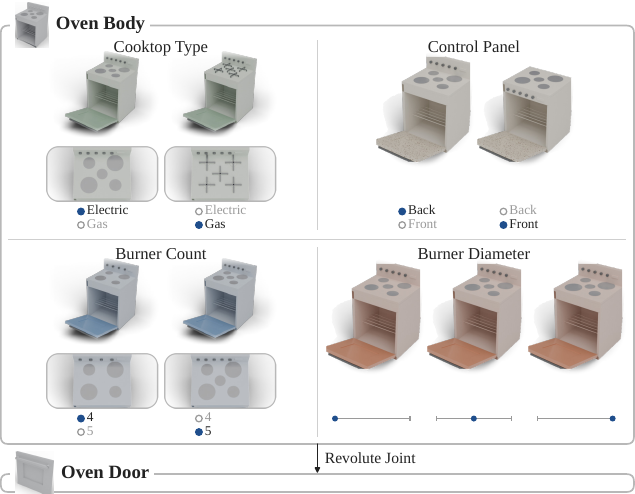}\\[-1mm]
        {\footnotesize (a) Major parameters of the \assetgen oven range generator}
    \end{minipage}\hfill
    \begin{minipage}[b]{0.3566\textwidth}
        \centering
        \includegraphics[width=\linewidth]{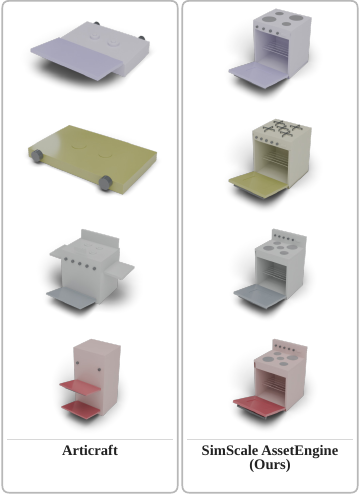}\\[-1mm]
        {\footnotesize (b) Qualitative comparison of different asset generation methods}
    \end{minipage}
    \caption{\textbf{Oven.} (a) The \assetgen generator varies the cooktop type, control panel, burner count, and burner size of the range. (b) Articraft produces a flat slab with a flap, a slab on wheels, a range with flaps sticking out of its sides, and a tall box with two doors.}
    \label{fig:assets-oven}
\end{figure*}

\begin{figure*}[!t]
    \centering
    \begin{minipage}[b]{0.5698\textwidth}
        \centering
        \includegraphics[width=\linewidth]{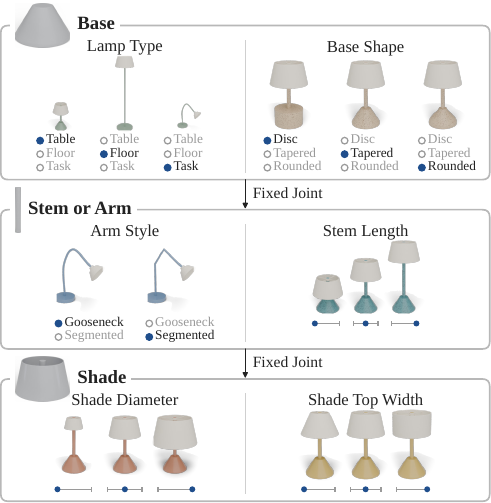}\\[-1mm]
        {\footnotesize (a) Major parameters of the \assetgen lamp generator}
    \end{minipage}\hfill
    \begin{minipage}[b]{0.4182\textwidth}
        \centering
        \includegraphics[width=\linewidth]{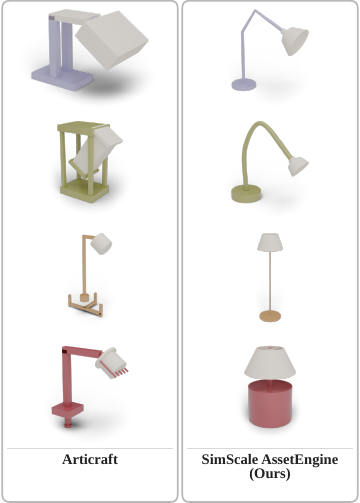}\\[-1mm]
        {\footnotesize (b) Qualitative comparison of different asset generation methods}
    \end{minipage}
    \caption{\textbf{Lamp.} (a) The \assetgen generator varies the type, base, arm, stem length, and shade of the lamp. (b) Articraft produces an oversized cube head (first), a box inside a cage (second), and a shade with rods attached along its outside and protruding past its rim (last).}
    \label{fig:assets-lamp}
\end{figure*}

\begin{figure*}[!t]
    \centering
    \begin{minipage}[b]{0.6768\textwidth}
        \centering
        \includegraphics[width=\linewidth]{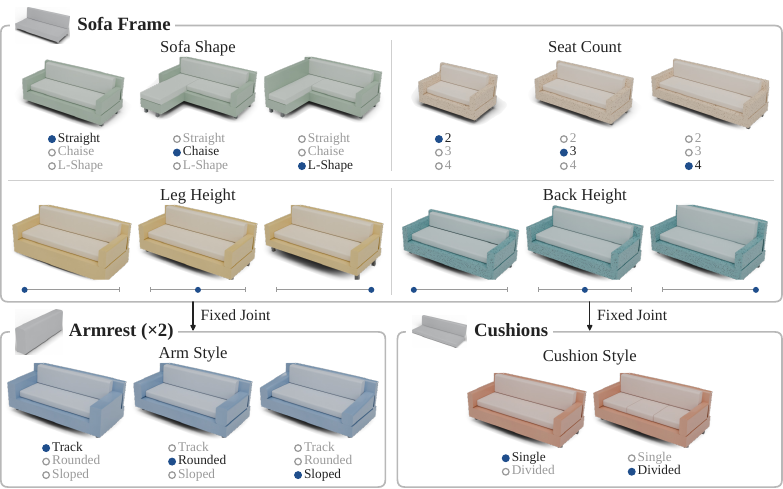}\\[-1mm]
        {\footnotesize (a) Major parameters of the \assetgen sofa generator}
    \end{minipage}\hfill
    \begin{minipage}[b]{0.3112\textwidth}
        \centering
        \includegraphics[width=\linewidth]{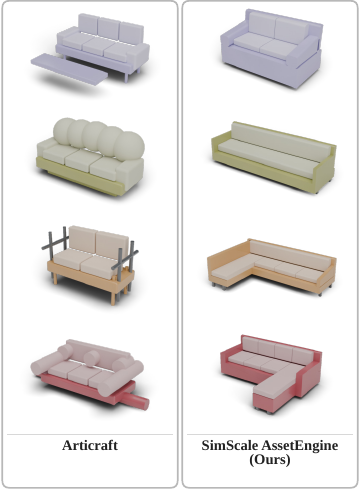}\\[-1mm]
        {\footnotesize (b) Qualitative comparison of different asset generation methods}
    \end{minipage}
    \caption{\textbf{Sofa.} (a) The \assetgen generator varies the shape, seat count, arms, cushions, leg height, and back height of the sofa. (b) The seat of the first Articraft sofa floats in front of it, the second has spheres as back cushions, the third has posts piercing its arms, and the last has cushions scattered over a slab.}
    \label{fig:assets-sofa}
\end{figure*}

\begin{figure*}[!t]
    \centering
    \begin{minipage}[b]{0.6689\textwidth}
        \centering
        \includegraphics[width=\linewidth]{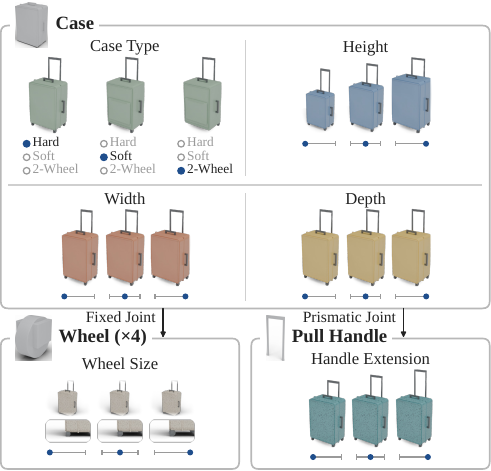}\\[-1mm]
        {\footnotesize (a) Major parameters of the \assetgen suitcase generator}
    \end{minipage}\hfill
    \begin{minipage}[b]{0.3191\textwidth}
        \centering
        \includegraphics[width=\linewidth]{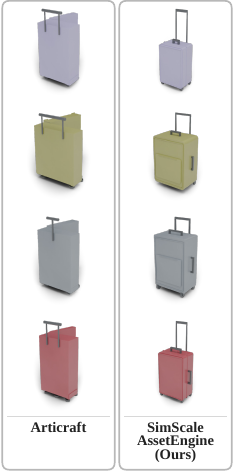}\\[-1mm]
        {\footnotesize (b) Qualitative comparison of different asset generation methods}
    \end{minipage}
    \caption{\textbf{Suitcase.} (a) The \assetgen generator varies the type, size, wheels, and handle extension of the suitcase. (b) Articraft lacks diversity, with suitcases that are roughly identical hard-shell designs.}
    \label{fig:assets-suitcase}
\end{figure*}

\begin{figure*}[!t]
    \centering
    \begin{minipage}[b]{0.6789\textwidth}
        \centering
        \includegraphics[width=\linewidth]{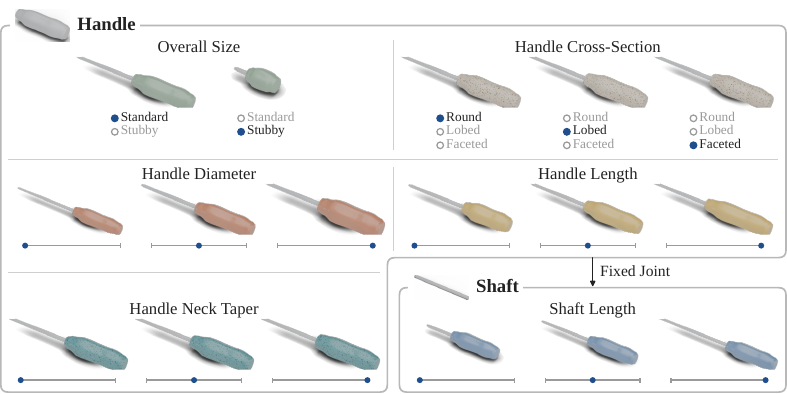}\\[-1mm]
        {\footnotesize (a) Major parameters of the \assetgen screwdriver generator}
    \end{minipage}\hfill
    \begin{minipage}[b]{0.3091\textwidth}
        \centering
        \includegraphics[width=\linewidth]{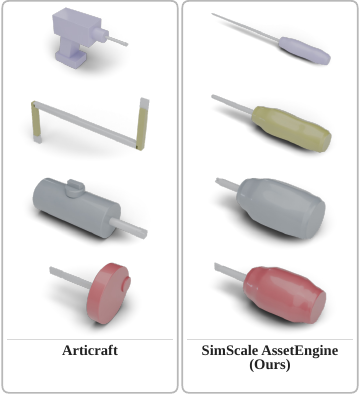}\\[-1mm]
        {\footnotesize (b) Qualitative comparison of different asset generation methods}
    \end{minipage}
    \caption{\textbf{Screwdriver.} (a) The \assetgen generator varies the size, handle section, and proportions of the screwdriver. (b) Articraft produces a boxy machine (first), a wide cylinder with a knob on top (third), and a shaft through a large disk (last), none of which resembles a typical screwdriver.}
    \label{fig:assets-screwdriver}
\end{figure*}

\begin{figure*}[!t]
    \centering
    \begin{minipage}[b]{0.6830\textwidth}
        \centering
        \includegraphics[width=\linewidth]{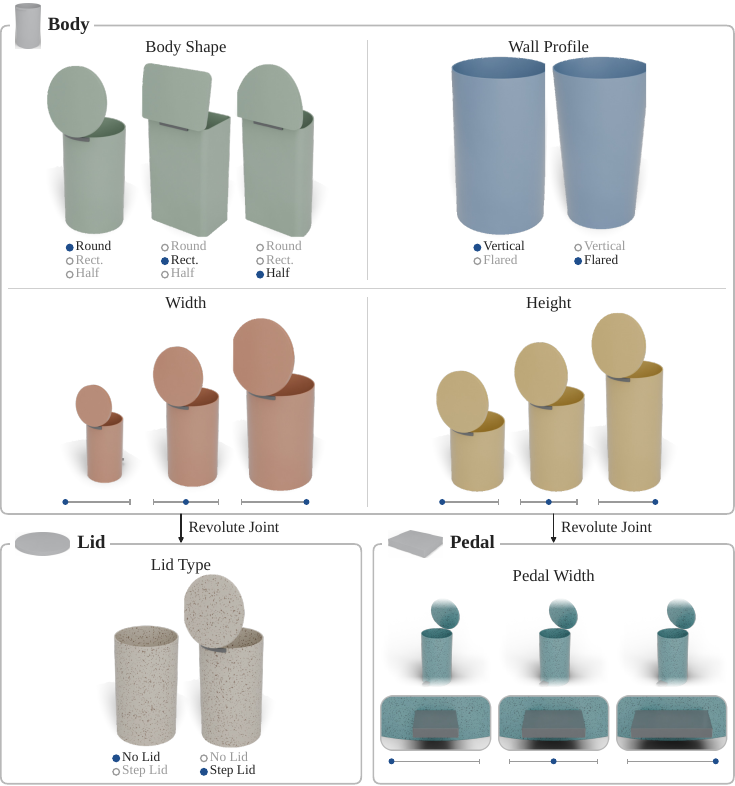}\\[-1mm]
        {\footnotesize (a) Major parameters of the \assetgen trash can generator}
    \end{minipage}\hfill
    \begin{minipage}[b]{0.3050\textwidth}
        \centering
        \includegraphics[width=\linewidth]{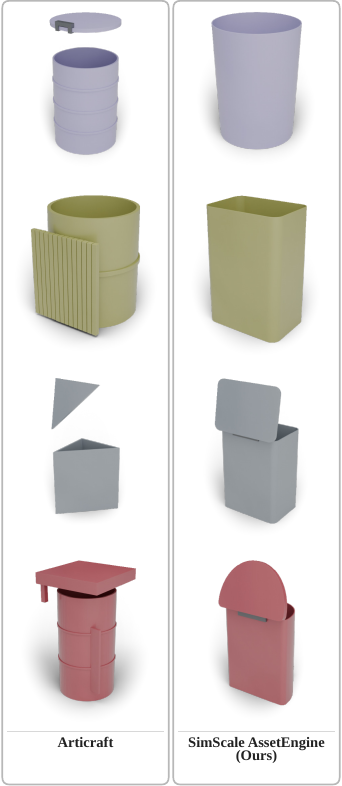}\\[-1mm]
        {\footnotesize (b) Qualitative comparison of different asset generation methods}
    \end{minipage}
    \caption{\textbf{Trash can.} (a) The \assetgen generator varies the body shape, wall profile, size, lid, and pedal of the trash can. (b) The lids of the first, third, and fourth Articraft cans are detached, and the lid of the fourth is also oversized. The third has an unrealistic triangular body, and the second has a slatted panel stuck to its side.}
    \label{fig:assets-trash-can}
\end{figure*}

\begin{figure*}[!t]
    \centering
    \begin{minipage}[b]{0.6789\textwidth}
        \centering
        \includegraphics[width=\linewidth]{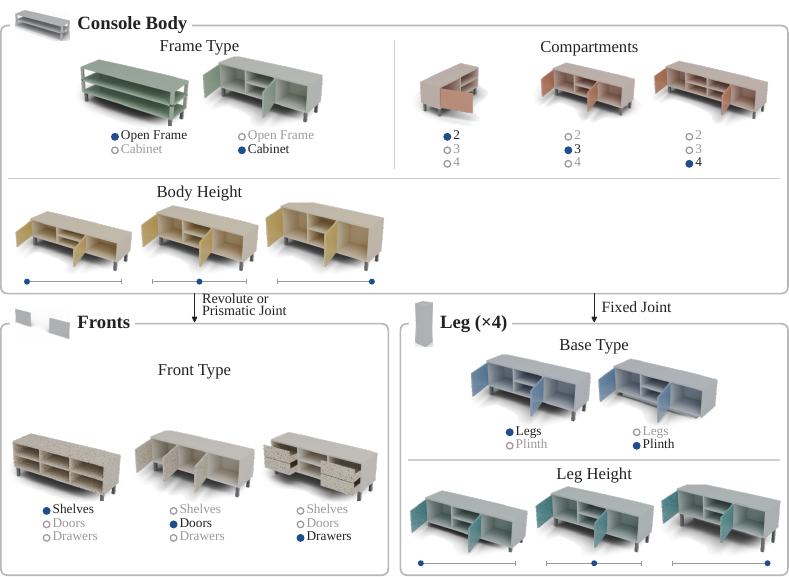}\\[-1mm]
        {\footnotesize (a) Major parameters of the \assetgen TV stand generator}
    \end{minipage}\hfill
    \begin{minipage}[b]{0.3091\textwidth}
        \centering
        \includegraphics[width=\linewidth]{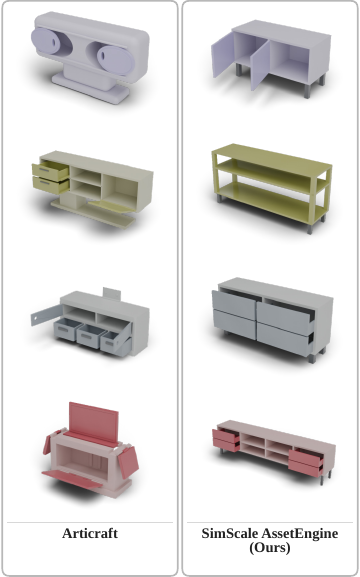}\\[-1mm]
        {\footnotesize (b) Qualitative comparison of different asset generation methods}
    \end{minipage}
    \caption{\textbf{TV stand.} (a) The \assetgen generator varies the frame, fronts, base, compartments, body height, and leg height of the TV stand. (b) The first Articraft sample does not look like a TV stand, and the last is a box with flaps opening on every side.}
    \label{fig:assets-tv-stand}
\end{figure*}

\begin{figure*}[!t]
    \centering
    \begin{minipage}[b]{0.7010\textwidth}
        \centering
        \includegraphics[width=\linewidth]{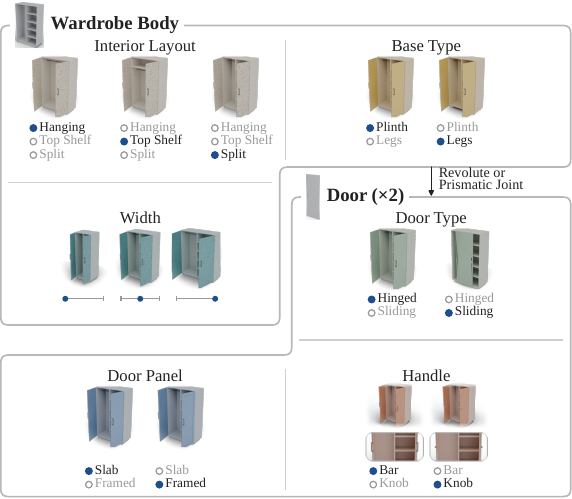}\\[-1mm]
        {\footnotesize (a) Major parameters of the \assetgen two-door wardrobe generator}
    \end{minipage}\hfill
    \begin{minipage}[b]{0.2870\textwidth}
        \centering
        \includegraphics[width=\linewidth]{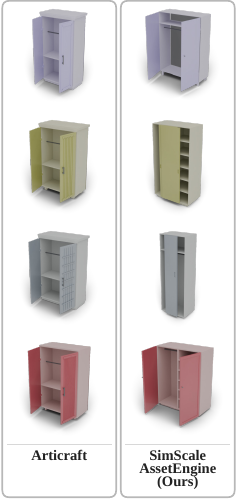}\\[-1mm]
        {\footnotesize (b) Qualitative comparison of different asset generation methods}
    \end{minipage}
    \caption{\textbf{Wardrobe.} (a) The \assetgen generator varies the doors, interior layout, door panels, handles, base, and width of the wardrobe. (b) Articraft lacks diversity, with wardrobes that share roughly the same overall shape and interior layout and vary only slightly in size.}
    \label{fig:assets-wardrobe}
\end{figure*}

\noindent\textbf{Survey design.}
Each survey page shows one category with one anonymous set of 40 assets per method, labeled Set A, B, or C in random order. Each set is shown in the default state and with all movable parts halfway open. The 5 task-relevant categories have sets from \system, Articraft, and SimFoundry cousins. The 15 additional categories (Figs.~\ref{fig:assets-bed}--\ref{fig:assets-wardrobe}) have sets from \system and Articraft only, since SimFoundry requires real-world videos.

\noindent\textbf{Criteria.}
For every category, participants rank the sets once per criterion. \emph{Geometric realism/plausibility} asks which set is most believable as typical real-world examples of the category, including coherent construction, connections, and articulation, considering every object in the set. \emph{Geometric diversity/coverage} asks which set best spans meaningful real-world geometric variation, where malformed, unrealistic, or out-of-category objects do not count as coverage.

\noindent\textbf{Participants.}
We collected 20 responses, which give 86 three-way rankings per criterion on the task-relevant categories and 270 two-way rankings per criterion on the additional categories.

\noindent\textbf{Metrics.}
In Fig.~\ref{fig:asset-preference-all-categories}a, each ranking contributes two top-2 selections, and the bars show the share of each method. In Fig.~\ref{fig:asset-preference-all-categories}b, the bars show the share of pairwise comparisons between \system and Articraft won by each method over all 20 categories.

\subsection{Asset Distribution Analysis}
\label{app:distributions}

Sec.~\ref{sec:exp_asset} compares asset generation methods through policy performance and human preference. Here we analyze the distributions of task-relevant assets in the parameter space of the \assetgen generators and in DINOv3~\cite{simeoni2025dinov3} feature space. For each asset, we find its nearest neighbor in the \assetgen generator (App.~\ref{app:cousins}) and normalize each parameter by its range, so that 0 and 1 mark the two ends of the range. For each category, we use 40 assets from each asset generation method to conduct the analysis.

\begin{figure*}[p]
{\centering
\includegraphics[width=\textwidth]{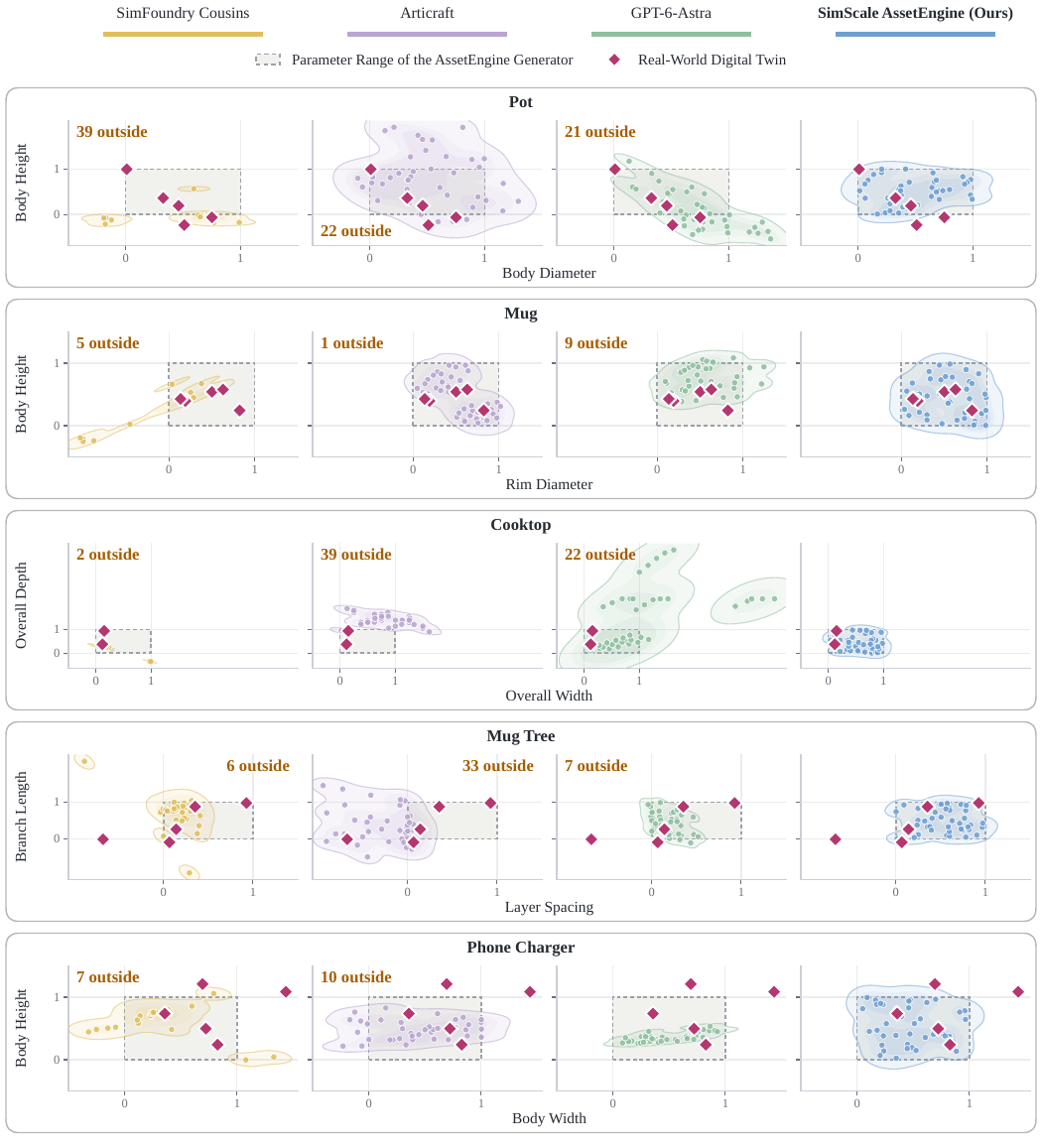}
\par}
\vspace{3pt}
{\footnotesize\floatcounter\label{fig:param-all}Fig.~\thefigure: \textbf{\assetgen assets evenly cover the generator's parameter ranges, while assets from other methods cluster in smaller regions or extend beyond these ranges.} Each row shows 40 assets per method on two major parameters, normalized so that the shaded square is the parameter range of the generator. Assets from other methods and real-world digital twins are projected into this parameter space. Dark-orange numbers count the assets outside it, and magenta diamonds mark real-world digital twins.\par}
\end{figure*}

\begin{figure*}[p]
{\centering
\includegraphics[width=0.97\textwidth]{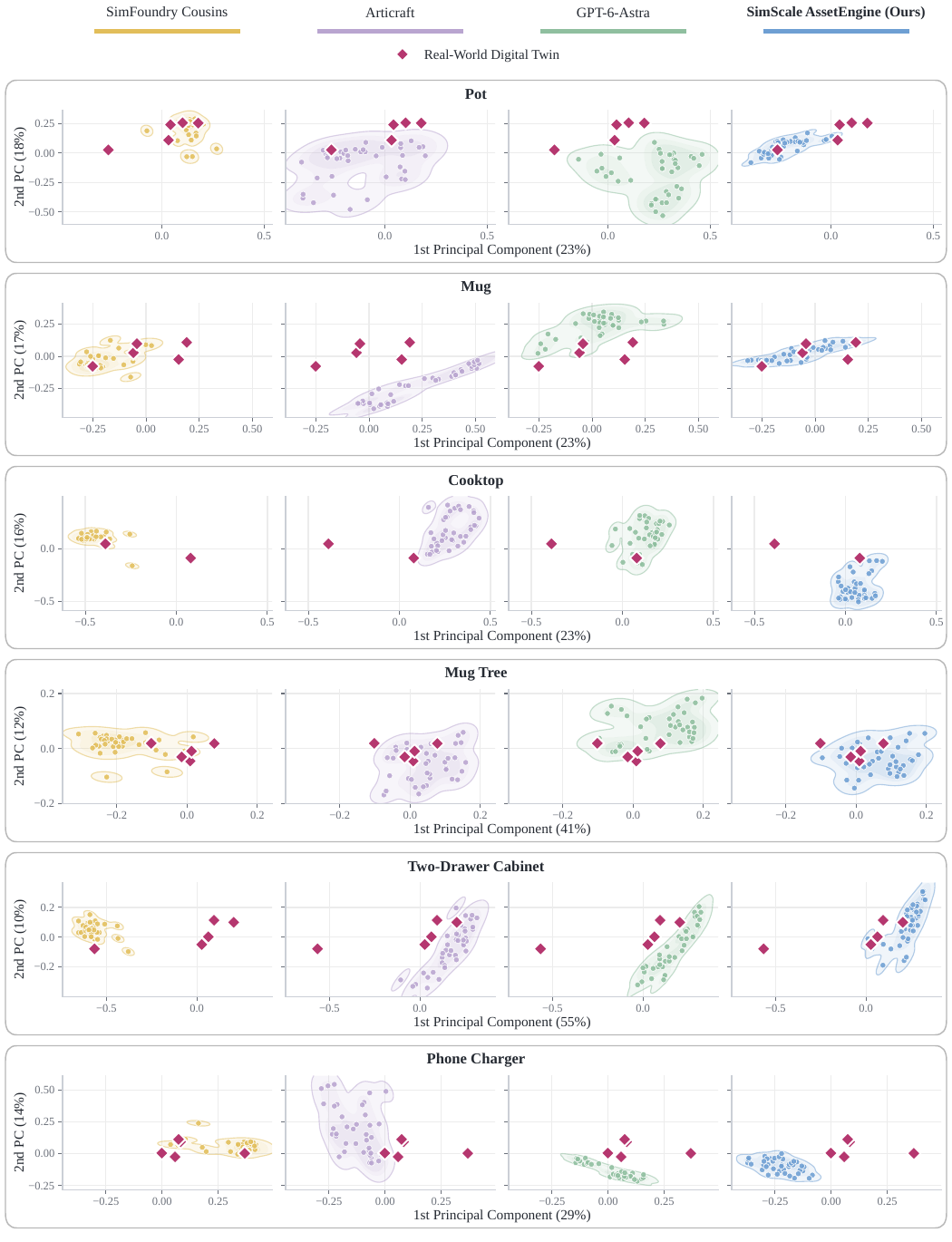}
\par}
\vspace{3pt}
{\footnotesize\floatcounter\label{fig:dino-all}Fig.~\thefigure: \textbf{Asset distributions in DINOv3 feature space.} Each row shows 40 assets per method on the first two principal components of DINOv3 embedding space. Magenta diamonds mark all 27 real-world twins: five per category except two cooktops.\par}
\end{figure*}

\noindent\textbf{\assetgen assets evenly cover the generator's parameter ranges, while assets of other methods either cluster in a small region or fall outside these ranges.}
Fig.~\ref{fig:param-all} shows the asset distributions in parameter space. SimFoundry cousins collapse into tight clusters, consistent with reconstructing all digital cousins of a category from one real-world object. Articraft assets spread more widely but leave structured gaps. For example, its mugs split into two disjoint groups, and 33 of its 40 mug trees fall outside the plotted parameter ranges. GPT-6-Astra assets also spread beyond the parameter ranges, with 21 of 40 pots and 22 of 40 cooktops outside them. In contrast, \assetgen assets fill the parameter ranges in every category. Fig.~\ref{fig:dino-all} also shows compact SimFoundry clusters in DINOv3 feature space.

\begin{figure*}[tp]
{\centering
\includegraphics[width=\textwidth]{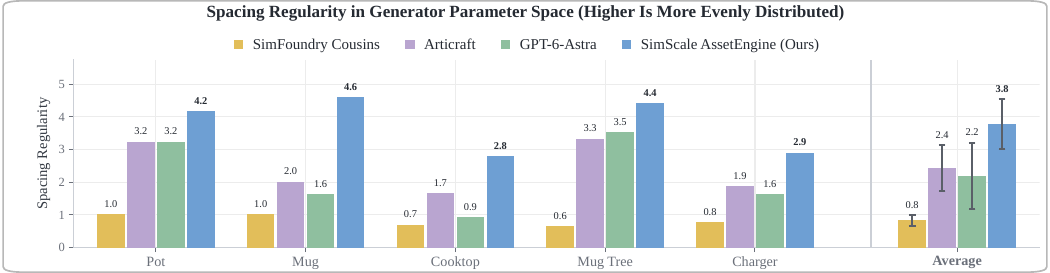}\\[-1mm]
{\footnotesize (a) Spacing regularity of assets in the parameter space of the \assetgen generators}\\[5mm]
\includegraphics[width=\textwidth]{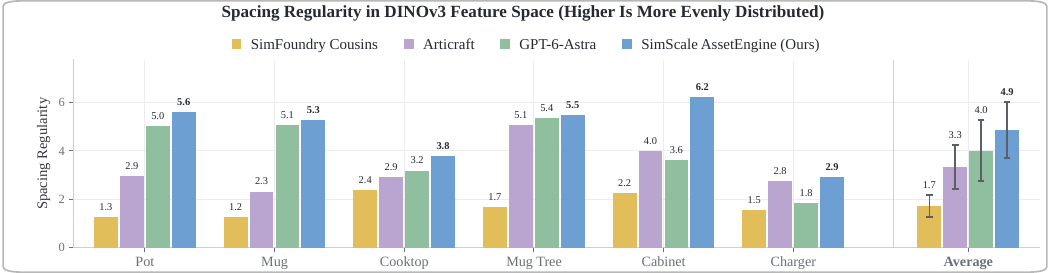}\\[-1mm]
{\footnotesize (b) Spacing regularity of assets in DINOv3 feature space}
\par}
\vspace{3pt}
{\footnotesize\floatcounter\label{fig:metrics-all}Fig.~\thefigure: \textbf{\assetgen assets are the most evenly distributed in both the generator parameter space and DINOv3 feature space.} Spacing regularity is the ratio of the mean to the standard deviation of the nearest-neighbor distances among the 40 assets of each method, using Euclidean distance between normalized parameters and between unit-normalized DINOv3 features.\par}
\end{figure*}

\noindent\textbf{\assetgen assets are the most evenly distributed in both the generator parameter space and DINOv3 feature space.}
We measure how evenly a method covers a space by its spacing regularity, which is high when assets are distributed evenly and low when they form clusters separated by gaps. As shown in Fig.~\ref{fig:metrics-all}, \assetgen assets are the most evenly distributed in all five categories in parameter space and in all six categories in DINOv3 feature space.

\begin{figure*}[tp]
\vspace{8pt}
{\centering
\includegraphics[width=\textwidth]{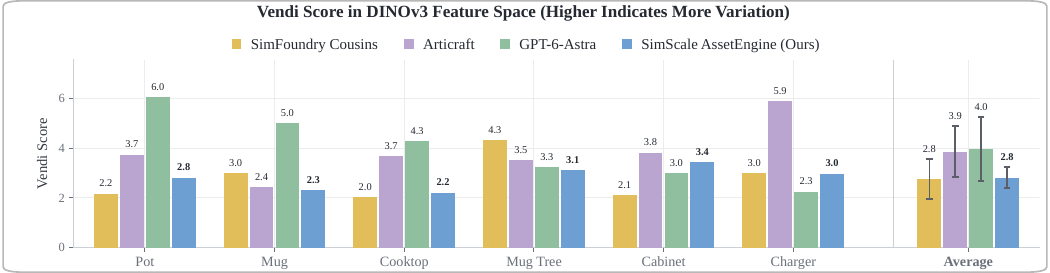}
\par}
\vspace{3pt}
{\footnotesize\floatcounter\label{fig:metrics-vendi}Fig.~\thefigure: \textbf{GPT-6-Astra and Articraft assets show greater DINOv3 feature variation on average, yet policies trained on them perform worse in the real world (Fig.~\ref{fig:real-world-assets}).} The Vendi score is higher when assets differ more from one another in DINOv3 feature space.\par}
\end{figure*}

\noindent\textbf{Greater visual variation alone does not lead to better policy performance.}
We further measure the visual variation of each method with the Vendi score~\cite{friedman2023vendi}, which is higher when the assets differ more from one another in DINOv3 feature space, without assessing geometric realism. Fig.~\ref{fig:metrics-vendi} shows the Vendi score of each method. GPT-6-Astra and Articraft have similar average Vendi scores, while \assetgen and SimFoundry have similar but lower scores. The user study instead defines diversity as geometric variation excluding unrealistic features (App.~\ref{app:user-study}). Participants rate \assetgen higher than SimFoundry in both geometric realism and diversity (Fig.~\ref{fig:asset-preference-all-categories}a). This contrast suggests that unrealistic or implausible objects contribute to the visual variation of VLM- and agent-based methods. Examples include the irregular bodies of GPT-6-Astra mugs, the baking dish and long-handled saucepan it generates as pots, and the wavy rim and double-loop handle of SimFoundry cousins (Fig.~\ref{fig:asset-qualitative} and App.~\ref{app:assets}).

However, policies trained on GPT-6-Astra and Articraft assets reach an average real-world policy performance of only 43.4\% and 36.6\% over the three tasks, against 84.9\% for \assetgen assets (Fig.~\ref{fig:real-world-assets}). Even with only 10 \assetgen assets and one human demonstration, \system reaches 57.6\% (Fig.~\ref{fig:real-world-scaling}), outperforming both methods with four times more assets. Training on unrealistic assets exposes the policy to object geometries far from the real-world distribution $p_{\mathrm{real}}$. In terms of the problem setup in Sec.~\ref{sec:problem-formulation}, the Vendi score measures feature-space variation in $p_{\mathrm{sim}}$, but not how much probability it places on physically implausible objects far away from $p_{\mathrm{real}}$. Such a $p_{\mathrm{sim}}$ can be visually varied yet train worse policies. This shows that increasing visual variation or the number of assets alone does not necessarily improve policy performance. A good $p_{\mathrm{sim}}$ must cover the geometric variations of $p_{\mathrm{real}}$ while placing little probability on physically implausible objects, which \assetgen achieves with category-level procedural generators.

\begin{figure*}[!t]
    \centering
    \begin{minipage}[b]{0.6773\textwidth}
        \centering
        \includegraphics[width=\linewidth]{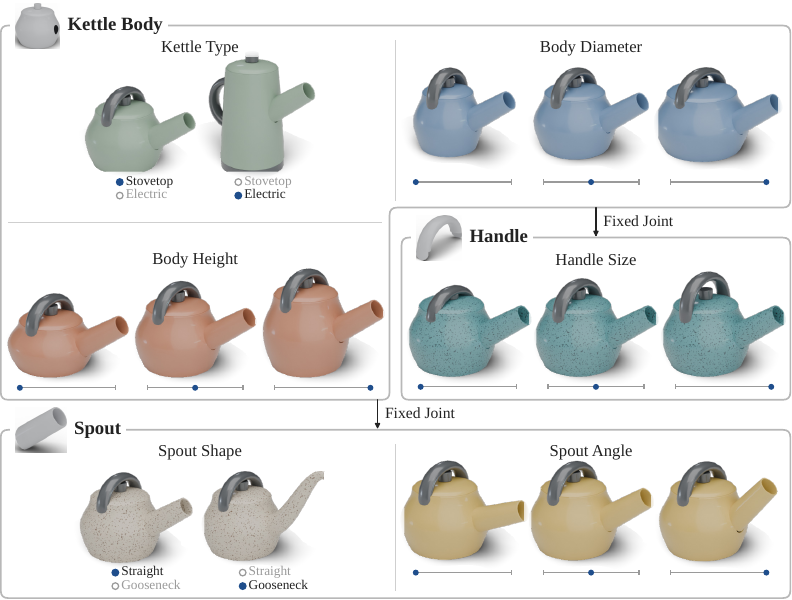}\\[-1mm]
        {\footnotesize (a) Major parameters of the \assetgen kettle generator}
    \end{minipage}\hfill
    \begin{minipage}[b]{0.3107\textwidth}
        \centering
        \includegraphics[width=\linewidth]{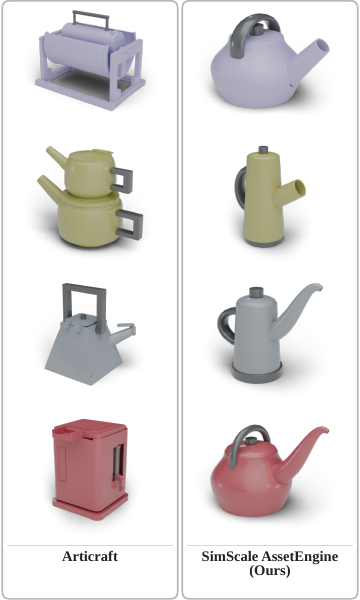}\\[-1mm]
        {\footnotesize (b) Qualitative comparison of different asset generation methods}
    \end{minipage}
    \caption{\textbf{Kettle.} (a) The \assetgen generator varies the type, spout, body size, and handle size of the kettle. (b) Articraft produces a horizontal cylinder in a box frame that does not resemble a kettle (first), two kettles unrealistically stacked together (second), and a box-shaped appliance that does not resemble a kettle (last).}
    \label{fig:assets-kettle}
\end{figure*}

\noindent\textbf{Articraft and GPT-6-Astra struggle to balance geometric diversity and realism.}
We observe that when asked for 40 assets of a category, Articraft first writes 40 asset-level prompts that are semantically different from one another, and then generates one asset from each prompt. While the first prompts describe common designs, later prompts drift toward rare or contrived designs to stay different from the earlier ones. For kettles, the first prompts ask for ``a classic polished stainless-steel stovetop whistling kettle with a broad dome body,'' while later prompts ask for ``a solar camping kettle designed around a cylindrical vacuum vessel in an articulated reflective trough cradle'' and ``a geometric pyramid-sided whistling stovetop kettle,'' which produce the implausible Articraft kettles in Fig.~\ref{fig:assets-kettle}. GPT-6-Astra proposes many fixed, diverse designs and varies their parameters widely to increase coverage. However, when prompted for a large number of assets, it still defines unrealistic designs that do not resemble typical objects in the target category. In contrast, an \assetgen generator uses parameter ranges and part relationships grounded in real-world references and refined through coverage review and geometric validation. This is consistent with the qualitative comparison in Sec.~\ref{sec:exp_asset}.

\begin{table*}[p]
\centering
\footnotesize
\begin{tblr}{width=\textwidth, colspec={Q[c,m,wd=0.11\textwidth] Q[c,m,wd=0.135\textwidth] X[j,m]}, rowsep=0.8pt, colsep=4pt}
\toprule
\textbf{Category} & \textbf{Method} & \textbf{Prompt} \\
\midrule
\SetCell[r=3]{c,m} Induction Cooktop & Articraft & Generate 40 geometrically different ordinary portable single-zone household induction cooktops. Exclude cookware, raised burners, protective rails, decorations, and obstructive controls. No need to texture objects \\ \cmidrule{2-3}
 & GPT-6-Astra & Please define 50 geometrically different induction cooktops. I only care about geometry, so no need to texture the assets. \\ \cmidrule{2-3}
 & \assetgen (ours) & Generate ordinary portable single-zone induction cooktops. Exclude cookware, raised burners, protective rails, decorations, and obstructive controls. \\
\midrule
\SetCell[r=3]{c,m} Cooking Pot / Stock Pot & Articraft & Generate 40 geometrically different ordinary household cooking pots/stock pots with exactly two handles. Exclude decorative forms and unrelated accessories. Feel free to search online for common geometries, handle types, and dimensions. No need to texture objects \\ \cmidrule{2-3}
 & GPT-6-Astra & Please define 50 geometrically different cooking pots with exactly two handles. I only care about geometry, so no need to texture them. \\ \cmidrule{2-3}
 & \assetgen (ours) & Generate ordinary cooking pots/stock pots with exactly two handles. Exclude decorative forms and unrelated accessories. \\
\midrule
\SetCell[r=3]{c,m} Mug / Teacup & Articraft & Generate 40 geometrically different ordinary household mugs and teacups with exactly one side handle. Exclude lids, saucers, decorations, and specialty forms. No need to texture the objects \\ \cmidrule{2-3}
 & GPT-6-Astra & Please generate 50 geometrically different mugs. I only care about geometry, so no need to texture them. \\ \cmidrule{2-3}
 & \assetgen (ours) & Generate ordinary mugs/teacups with exactly one side handle. Exclude lids, saucers, and decorations. \\
\midrule
\SetCell[r=3]{c,m} Mug Tree & Articraft & Generate 40 geometrically different ordinary household mug trees with exactly three branch layers and exactly two opposed branches per layer, with adjacent layers rotated by 90 degrees. Exclude extra hooks, obstacles, and decorations. No need to texture the objects \\ \cmidrule{2-3}
 & GPT-6-Astra & Please generate 50 geometrically different mug trees. Each mug tree must have 3 layers of branches. Each layer has exactly 2 branches, separated by 180 degrees. Adjacent layers should be 90 degrees off in terms of branch orientation. I only care about geometry, so no need to texture assets. \\ \cmidrule{2-3}
 & \assetgen (ours) & Generate an ordinary mug tree. It must have exactly three branch layers and exactly two opposite branches per layer, producing six branches total. The two branches within each layer must be separated by 180 degrees, and each adjacent layer must be rotated by 90 degrees. Exclude decorations and other obstacles. Feel free to search online to determine the dimensions of the central post and branches. \\
\midrule
\SetCell[r=3]{c,m} Two-Drawer Cabinet & Articraft & Generate 40 geometrically different household two-drawer cabinets/file cabinets/nightstands. Every instance must have two identical, vertically stacked drawers. Each drawer must have a pull/arc handle. Avoid knobs. For overall size, I want left-right to be within 30-60cm, front-back to be within 25-45cm, and height to be within 30-70cm. \\ \cmidrule{2-3}
 & GPT-6-Astra & Generate 50 two-drawer cabinets, file cabinets, or nightstands. Every instance must have two identical, vertically stacked drawers. Each drawer must have a pull/arc handle. Avoid knobs. For overall size, I want each instance left-right to be within 30-60cm, front-back to be 25-45cm, and height to be 30-70cm. Each object must have its articulation labeled. \\ \cmidrule{2-3}
 & \assetgen (ours) & Generate a procedural generator of common two-drawer cabinets, file cabinets, and nightstands. Every instance must have two identical, vertically stacked drawers. Each drawer must have a pull/arc handle. Avoid knobs. For size, I want left-right to be 30-60cm, front-back to be 25-45cm, and height to be 30-70cm. \\
\midrule
\SetCell[r=3]{c,m} Phone Charger / Wall Power Adapter & Articraft & Please generate 40 geometrically different phone chargers/wall power adapters. No need to texture objects \\ \cmidrule{2-3}
 & GPT-6-Astra & Generate 50 phone chargers/wall power adapters. \\ \cmidrule{2-3}
 & \assetgen (ours) & Generate a procedural generator for phone chargers and wall power adapters. \\
\bottomrule
\end{tblr}
\par\vspace{3pt}
{\footnotesize\floatcounter\label{tab:prompts-task}TABLE~\thetable: Prompts of each method for the task-relevant categories.\par}
\vspace{-4mm}
\end{table*}

\begin{table*}[p]
\centering
\footnotesize
\begin{tblr}{width=\textwidth, colspec={Q[c,m,wd=0.11\textwidth] Q[c,m,wd=0.135\textwidth] X[j,m]}, rowsep=0.8pt, colsep=4pt}
\toprule
\textbf{Category} & \textbf{Method} & \textbf{Prompt} \\
\midrule
\SetCell[r=2]{c,m} Bed & Articraft & Generate 40 geometrically different beds. No need to texture \\ \cmidrule{2-3}
 & \assetgen (ours) & Generate a procedural generator for bed. \\
\midrule
\SetCell[r=2]{c,m} Bottle & Articraft & Generate 40 geometrically different bottles. No need to texture \\ \cmidrule{2-3}
 & \assetgen (ours) & bottle \\
\midrule
\SetCell[r=2]{c,m} Cabinet & Articraft & Generate 40 geometrically different cabinets. No need to texture \\ \cmidrule{2-3}
 & \assetgen (ours) & cabinet \\
\midrule
\SetCell[r=2]{c,m} Chair & Articraft & Generate 40 geometrically different chairs. No need to texture \\ \cmidrule{2-3}
 & \assetgen (ours) & Generate a procedural generator for chair. \\
\midrule
\SetCell[r=2]{c,m} Kettle & Articraft & Generate 40 geometrically different kettles. No need to texture \\ \cmidrule{2-3}
 & \assetgen (ours) & A category-level procedural generator for ordinary kettles. \\
\midrule
\SetCell[r=2]{c,m} Lamp & Articraft & Generate 40 geometrically different lamps. No need to texture \\ \cmidrule{2-3}
 & \assetgen (ours) & Generate a procedural generator for Lamp. \\
\midrule
\SetCell[r=2]{c,m} Oven / stove & Articraft & Generate 40 geometrically different ovens / stoves. No need to texture \\ \cmidrule{2-3}
 & \assetgen (ours) & Generate a procedural generator for Oven / stove \\
\midrule
\SetCell[r=2]{c,m} Refrigerator & Articraft & Generate 40 geometrically different refrigerators. No need to texture \\ \cmidrule{2-3}
 & \assetgen (ours) & Generate a procedural generator for Refrigerator \\
\midrule
\SetCell[r=2]{c,m} Screwdriver & Articraft & Generate 40 geometrically different screwdrivers. No need to texture \\ \cmidrule{2-3}
 & \assetgen (ours) & A category-level procedural generator for ordinary handheld screwdrivers. \\
\midrule
\SetCell[r=2]{c,m} Sofa & Articraft & Generate 40 geometrically different sofas. No need to texture \\ \cmidrule{2-3}
 & \assetgen (ours) & Generate a procedural generator for sofa. \\
\midrule
\SetCell[r=2]{c,m} Suitcase & Articraft & Generate 40 geometrically different suitcases. No need to texture \\ \cmidrule{2-3}
 & \assetgen (ours) & A procedural generator for common suitcases. \\
\midrule
\SetCell[r=2]{c,m} Table & Articraft & Generate 40 geometrically different tables. No need to texture \\ \cmidrule{2-3}
 & \assetgen (ours) & table \\
\midrule
\SetCell[r=2]{c,m} Trash can & Articraft & Generate 40 geometrically different trash cans. No need to texture \\ \cmidrule{2-3}
 & \assetgen (ours) & Trash can \\
\midrule
\SetCell[r=2]{c,m} TV stand / media console & Articraft & Generate 40 geometrically different TV stands / media consoles. No need to texture \\ \cmidrule{2-3}
 & \assetgen (ours) & Generate a procedural generator for TV stand / media console. \\
\midrule
\SetCell[r=2]{c,m} Two-door wardrobe & Articraft & Generate 40 geometrically different two-door wardrobes. Each wardrobe must have two identical doors stacked horizontally \\ \cmidrule{2-3}
 & \assetgen (ours) & Ordinary two-door wardrobe. Each wardrobe must have two identical doors stacked horizontally. \\
\bottomrule
\end{tblr}
\par\vspace{3pt}
{\footnotesize\floatcounter\label{tab:prompts-additional}TABLE~\thetable: Prompts of each method for the 15 additional categories of the user study.\par}
\end{table*}

\end{document}